\documentclass[twoside,11pt]{article}

\usepackage{blindtext}

\usepackage[preprint]{jmlr2e}
\usepackage[aboveskip=\abovecaptionskip,belowskip=\belowcaptionskip]{caption}

\usepackage{amssymb}            
\usepackage{mathtools}          
\usepackage{mathrsfs}           
\usepackage{graphicx}           
\usepackage[space]{grffile}     
\usepackage{url}                
\usepackage{lipsum}             

\usepackage{orcidlink}
\usepackage[utf8]{inputenc} 
\usepackage[T1]{fontenc}    
\usepackage{hyperref}       
\usepackage{url}            
\usepackage{booktabs}       
\usepackage{amsfonts}       
\usepackage{nicefrac}       
\usepackage{microtype}      
\usepackage{makecell}
\usepackage{cleveref}
\usepackage{graphicx}

\usepackage{wrapfig}

\usepackage{amssymb}
\usepackage{bm}
\usepackage{placeins}
\usepackage{appendix}
\usepackage{multirow}
\definecolor{dark-blue}{rgb}{0,0,0.7}
\hypersetup{
    colorlinks, linkcolor={dark-blue},
    citecolor={dark-blue}, urlcolor={dark-blue}
}
\newcommand{\blackhref}[3][black]{\href{#2}{\color{#1}{#3}}}%
\newcommand{\norm}[1]{\left\lVert#1\right\rVert}

\usepackage{lastpage}

\ShortHeadings{Tackling NCMDPs using Reinforcement Learning}{N\"agele, F\"osel, Olle, Zen and Marquardt}
\firstpageno{1}

\begin{document}

\title{Tackling Decision Processes with Non-Cumulative\\Objectives using Reinforcement Learning}

\author{\name \blackhref{https://orcid.org/0000-0001-6382-2077}{Maximilian N\"agele$^{1,2}$} \email maximilian.naegele@mpl.mpg.de \\
       \name \blackhref{https://orcid.org/0000-0003-3338-5130}{Jan Olle$^{1}$} \\
       \name Thomas F\"osel$^{2}$ \\
       \name \blackhref{https://orcid.org/0000-0002-7645-125X}{Remmy Zen$^{1}$}\\
       \name \blackhref{https://orcid.org/0000-0003-4566-1753}{Florian Marquardt$^{1,2}$}\\
       \addr
       $^{1}$Max Planck Institute for the Science of Light, Germany\\ $^{2}$Friedrich-Alexander-Universit\"at Erlangen-N\"urnberg
       }
\editor{My editor}

\maketitle

\begin{abstract}%
Markov decision processes (MDPs) are used to model a wide variety of applications ranging from game playing over robotics to finance. Their optimal policy typically maximizes the expected sum of rewards given at each step of the decision process. However, many real-world problems do not fit straightforwardly into this framework:  Non-cumulative Markov decision processes (NCMDPs), where instead of the expected sum of rewards, the expected value of an arbitrary function of the rewards is maximized. Example functions include the maximum of the rewards or their mean divided by their standard deviation. In this work, we introduce a general mapping of NCMDPs to standard MDPs. This allows all techniques developed to find optimal policies for MDPs, such as reinforcement learning or dynamic programming, to be directly applied to the larger class of NCMDPs. We demonstrate the effectiveness of our approach in diverse reinforcement learning tasks, including classical control, financial portfolio optimization, and discrete optimization. Our approach improves both final performance and training efficiency compared to relying on standard MDPs.
\end{abstract}


\section{Introduction}\label{sec:introduction}
Markov decision processes (MDPs) are used to model a wide range of applications where an agent iteratively interacts with an environment during a trajectory.
Important examples include robotic control \citep{OpenAI2020}, game playing \citep{Mnih2015}, or discovering algorithms \citep{Mankowitz2023}.
At each time step $t$ of the MDP, the agent chooses an action based on the state of the environment and receives a reward $r_t$.
The agent's goal is to follow an ideal policy that maximizes 
\begin{equation}\label{eq:expect_sum}
    \mathbb{E}_\pi\left[\sum_{t=0}^{T-1}{r_t}\right],
\end{equation}
where $T$ is the length of the trajectory and the expectation value is taken over both the agent's probabilistic policy $\pi$ and the probabilistic environment.
There exist a multitude of established strategies for finding (approximately) ideal policies of MDPs, such as dynamic programming and reinforcement learning \citep{sutton1998introduction}.

A limitation of the framework of MDPs is the restriction to ideal policies that maximize \Cref{eq:expect_sum}, while a large class of problems cannot straightforwardly be formulated this way. For example, in weakest-link problems, the goal is to maximize the minimum rather than the sum of rewards, e.g.\ in network routing one wants to maximize the minimum bandwidths along a path \citep{cui2023}. In finance, the Sharpe ratio, i.e.\ the mean divided by the standard deviation of portfolio gains, is an important figure of merit of an investment strategy, since maximizing it will yield more risk-averse strategies than maximizing the sum of portfolio gains \citep{sharpe1966}.
We therefore require a method to tackle non-cumulative Markov decision processes (NCMDPs), where instead of the expected sum of rewards, the expectation value of an arbitrary function of the rewards is maximized.
The main contribution of this paper is twofold:

\begin{figure}[t]
    \centering
    \includegraphics{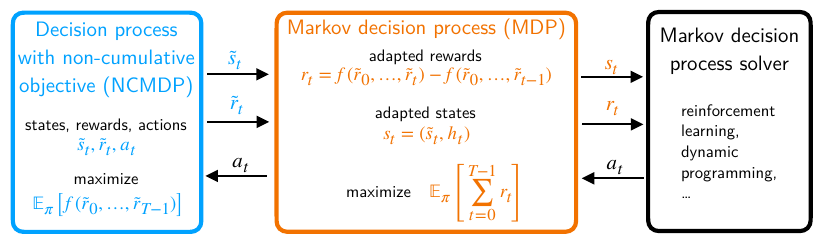}
    \caption{Mapping of a decision process with non-cumulative objective (blue) to a corresponding Markov decision process with adapted states and rewards (orange) that can be solved by standard methods (black). For details and notation, see \Cref{sec:theory}.}
    \label{fig:fig1}
\end{figure}

\begin{enumerate}
\item We expand the scope of reinforcement learning by introducing a theoretical framework that maps NCMDPs to standard MDPs. This mapping enables direct application of advanced MDP solvers, including reinforcement learning and dynamic programming, to NCMDPs without modification. Our approach accommodates general non-cumulative objectives and is straightforward to implement. \newline
\textbf{Context:} Existing methods for solving NCMDPs are restricted to either maximizing the highest reward \citep{quah2006, veviurko2024} or handling a limited set of objectives using Q-learning in deterministic settings \citep{cui2023}.
In contrast, our framework generalizes to arbitrary non-cumulative objectives in both deterministic and stochastic environments, albeit at the cost of increased state space size. In practice, this is not a fundamental limitation for deep reinforcement learning techniques, since they are well equipped to handle even exponentially large state spaces. Moreover, unlike prior approaches that require modifying the reinforcement learning algorithm, our method treats both the environment and learning algorithm as black boxes, simplifying implementation.
\item We conduct numerical experiments demonstrating the effectiveness of deep reinforcement learning on NCMDPs across complex applications with large state and action spaces, including portfolio optimization, classical control, and discrete optimization. Our results indicate improvements in training efficiency and final performance compared to prior reinforcement learning approaches. \newline
\textbf{Context:} The problems we investigate involve non-cumulative objectives and stochastic environments that place them beyond the scope of previous NCMDP-solving methods. As a result, prior state-of-the-art reinforcement learning approaches were constrained by the need to approximate non-cumulative objectives using standard MDPs. Our method overcomes this limitation by allowing direct specification of the exact objective to the reinforcement learning agent, leading to improved performance.
\end{enumerate}

\section{Theoretical Analysis}\label{sec:theory}

\subsection{Preliminaries}
In an MDP, an agent iteratively interacts with an environment during a trajectory. At each time step $t$, the agent receives the current state $s_t$ of the environment and selects an action $a_t$ by sampling from its policy $\pi(a_t |s_t)$, which is a probability distribution over all possible actions given a state. The next state of the environment $s_{t+1}$ and the agent's immediate reward $r_t$ are then sampled from the transition probability distribution $p(r_t, s_{t+1}| s_t, a_t)$, which depends on the MDP. This process repeats until it reaches a terminal state. The ideal policy of the agent maximizes the expected sum of immediate rewards, i.e.\ \Cref{eq:expect_sum}. Note that we take this ideal policy to be part of the definition of an MDP to distinguish MDPs from NCMDPs with different ideal policies. For simplicity, we discuss only non-discounted, episodic MDPs. However, generalization to discounted settings is straightforward.

\subsection{Non-cumulative Markov Decision Processes}
In the following, we denote states and rewards related to NCMDPs by $\tilde s_t$ and  $\tilde r_t$, respectively, to distinguish them from MDPs.
We define NCMDPs equivalently to MDPs except for their ideal policies maximizing the expectation value of an arbitrary scalar function $f$ of the immediate rewards $\tilde r_t$ instead of their sum, i.e.\
\begin{equation}\label{eq:max_f}
    \mathbb{E}_\pi\left[f(\tilde r_0, \tilde r_1, \dots, \tilde r_{T-1})\right].
\end{equation}
To accommodate trajectories of arbitrary finite length, we require $f$ to be a family of functions consisting of a function $f_t : \mathbb{R}^t \rightarrow \mathbb{R}$ for each $t \in \mathbb{N}$. For brevity, we denote $f_t(\tilde r_0, \tilde r_1, \dots, \tilde r_{t-1}) = f(\tilde r_0, \tilde r_1, \dots, \tilde r_{t-1})$.
NCMDPs still have a Markovian transition probability distribution but their return, i.e. \Cref{eq:max_f}, can depend on the rewards in a non-cumulative and therefore non-Markovian way. Thus, NCMDPs are a generalization of MDPs.
Due to this distinction, MDP solvers cannot straightforwardly be used for NCMDPs. To solve this problem, we describe a general mapping from an NCMDP $\tilde M$ to a corresponding MDP $M$ with adapted states and adapted rewards but the same actions. The mapping is chosen such that the optimal policy of $\tilde M$ is equivalent to the optimal policy of $M$. Therefore, a solution for $\tilde M$ can readily be obtained by solving $M$ with existing MDP solvers, as shown in \Cref{fig:fig1}. In the following, we formalize these ideas.

\begin{definition}\label{def:corr_MDP}
Given an NCMDP $\tilde M$ with rewards $\tilde r_t$, states $\tilde s_t$, actions $ a_t$, the Markovian state transition probability distribution $\tilde p(\tilde r_t, \tilde s_{t+1}| \tilde s_t, a_t)$, and non-cumulative objective function $f$, we define a \emph{corresponding} MDP $M$ with rewards $r_t$, states $s_t$, the same actions $a_t$, and state transition probability distribution $p(r_t, s_{t+1}| s_t, a_t)$ with functions $\rho$ and $u$, and vectors $h_t$ with initial values $h_0$ such that
\begin{equation}\label{eq:r_def}
   r_t = f(\tilde r_0,\dots, \tilde r_{t})  - f(\tilde r_0,\dots, \tilde r_{t-1}) = \rho(h_{t}, \tilde r_t),
\end{equation}
\begin{equation}\label{eq:s_def}
   s_t = (\tilde s_t, h_t) \text{ with } h_{t+1}=u(h_{t}, \tilde r_t), 
\end{equation}
\begin{equation}\label{eq:p_def}
    p\big(r_t, s_{t+1}{=}(\tilde s_{t+1}, h_{t+1})| s_t{=}(\tilde s_{t}, h_{t}), a_t\big) = \sum_{\tilde r_t} \tilde p(\tilde r_t, \tilde s_{t+1}| \tilde s_t,  a_t) \delta_{h_{t+1},u(h_{t}, \tilde r_t)} \delta_{r_t,\rho(h_{t}, \tilde r_t)},
\end{equation}
where $\delta$ is the Kronecker delta. For continuous probability distributions, the sum should be replaced by an integral and the $\delta$ by Dirac delta functions.
\end{definition}

We now provide some intuition for this definition: \Cref{eq:r_def} ensures that the return of $M$, i.e.\ $\sum_{t=0}^{T-1} r_t$, is equal to the return of $\tilde M$, i.e.\ $f(\tilde r_0,\dots, \tilde r_{T-1})$. To compute the immediate rewards~$r_t$ of $M$ in a purely Markovian manner, we need access to information about the previous rewards of the trajectory. This can be achieved by extending the state space with $h_t$, which preserves all necessary information about the reward history. The function $u$ in \Cref{eq:s_def} updates this information at each time step.
Finally, the Kronecker deltas in \Cref{eq:p_def} `pick' all the possible $\tilde r_t$ that result in the same $h_{t+1}$ and $r_t$. For example, for $f(\tilde r_0,\dots, \tilde r_t) = \min(\tilde r_0,\dots, \tilde r_t)$, we can choose $r_0 = h_1 =\tilde r_0$ followed by $h_{t+1} = u(h_{t}, \tilde r_t) = \min(h_{t}, \tilde r_t)$, and $r_t = \rho(h_{t}, \tilde r_t) = \min(0, \tilde r_t - h_{t})$. More examples are shown in \Cref{tab:ftable}. 

Due to \Cref{def:corr_MDP}, there is a mapping between trajectories of NCMDPs and corresponding MDPs, which we will exploit to use the same policy for both decision processes:

\begin{definition}\label{def:T_map}
    A trajectory $\mathcal{\tilde T} =  (\tilde s_0, a_0, \tilde r_0, \dots)$ of an NCMDP $\tilde M$ is \emph{mapped} onto a trajectory $\mathcal{T}  = \mathrm{map}\left(\mathcal{\tilde T}\right)= (s_0, a_0, r_0, \dots)$ of a corresponding MDP $M$ by calculating $r_t$ according to \Cref{eq:r_def}, setting $s_t = (\tilde s_t, h_t)$ with $h_t$ calculated according to \Cref{eq:s_def},  and keeping the actions $a_t$ the same.
\end{definition}
Note that multiple $\mathcal{\tilde T}$ potentially map to the same $\mathcal{T}$ since $\rho$ and $u$, and, therefore, the map operation are not necessarily injective. We can now state the main result of this manuscript: 

\setlength{\tabcolsep}{5pt}
\begin{table}[t]
\centering
\begin{tabular}{ccc}
 \toprule
 $\boldsymbol{f(\tilde r_0}, \dots , \boldsymbol{\tilde r_{t})}$ & Additional state information $\boldsymbol{h_t}$ & Adapted reward $\boldsymbol{r_t}$  \\ 
  \toprule
  $\max(\tilde r_0,\dots,\tilde r_{t})$ & $h_0=-\infty$, $h_{t+1} = \max(h_t, \tilde r_t)$ & \begin{tabular}{@{}c@{}}
  $r_0 = \tilde r_0$ \\
  $r_t = \max(0, \tilde r_t - h_t), t>0$ \end{tabular}  \\ 
   \midrule
  $\min(\tilde r_0,\dots,\tilde r_{t})$ & $h_0=\infty$, $h_{t+1} = \min(h_t, \tilde r_t)$ &\begin{tabular}{@{}c@{}}
  $r_0 = \tilde r_0$\\
  $r_t = \min(0, \tilde r_t - h_t), t>0$ \end{tabular}\\ 
  \midrule
 \makecell[c]{Sharpe ratio\\  $\frac{\mathrm{MEAN}(\tilde r_0,\dots,\tilde r_{t})}{\mathrm{STD}(\tilde r_0,\dots,\tilde r_{t})}$} & $h_0=\begin{bmatrix} 0\\  0\\  0 \end{bmatrix}$, $h_{t+1}=\begin{bmatrix} \frac{h_t^{(2)}}{h_t^{(2)}+1} h_t^{(0)} + \frac{1}{h_t^{(2)}+1} \tilde r_t\\  \frac{h_t^{(2)}}{h_t^{(2)}+1} h_t^{(1)} + \frac{1}{h_t^{(2)}+1} \tilde r_t^2\\\  h_t^{(2)} + 1 \end{bmatrix}$
& \begin{tabular}{@{}r@{}}$r_t = \frac{h_{t+1}^{(0)}}{\sqrt{h_{t+1}^{(1)}  - h_{t+1}^{(0)^2}}}$\\ $- \frac{h_{t}^{(0)}}{\sqrt{h_{t}^{(1)}  - h_{t}^{(0)^2}}} $\end{tabular}  \\  
 \midrule
 $\underset{k\in[-1, t]}{\max}\sum_{i=0}^k \tilde r_i$ & $h_0=0$, $h_{t+1} = \max(0, h_t - \tilde r_t)$ & $r_t$ = $\max(0, \tilde r_t - h_t)$  \\  
 \midrule
 $\tilde r_0 \tilde r_1 \dots \tilde r_{t}$ & $h_0 = 1$, $h_{t+1} = \tilde r_t h_t$ & \begin{tabular}{@{}c@{}}$r_0 = \tilde r_0,$ \\ $r_t = h_{t+1} - h_{t}, t>0$\end{tabular} \\ 
 \midrule
 \makecell[c]{Harmonic mean \\ $\frac{1}{\frac{1}{\tilde r_0} + \dots + \frac{1}{\tilde r_{t}}}$} & $h_0 = 0, h_{t+1} = h_t + \frac{1}{\tilde r_{t}}$ & \begin{tabular}{@{}c@{}}$r_0 = \tilde r_0,$ \\ $r_t = \frac{1}{h_{t+1}} - \frac{1}{h_t}, t>0$\end{tabular} \\ 
 \midrule
 \begin{tabular}{@{}c@{}} $\delta^{t} \sum_{k=0}^{t} \tilde r_k$, \\ $ \delta \in (0,1)$\end{tabular} & $h_{0} = \begin{bmatrix} 0 \\ 0 \end{bmatrix}$,  $h_{t+1} = \begin{bmatrix} h_{t}^{(0)} + \tilde r_t \\ h_{t}^{(1)} + 1 \end{bmatrix}$ & \begin{tabular}{@{}r@{}}$r_t = \delta^{h_{t+1}^{(1)}} h_{t+1}^{(0)}$ \\ $ -   \delta^{h_{t}^{(1)}} h_{t}^{(0)}$\end{tabular}\\
 \midrule
 $\frac{1}{t+1} \sum_{k=0}^{t} \tilde r_k$ & $h_{0} = \begin{bmatrix} 0 \\ 0 \end{bmatrix}$,  $h_{t+1} = \begin{bmatrix} h_{t}^{(0)} + \tilde r_t \\ h_{t}^{(1)} + 1 \end{bmatrix}$ & \begin{tabular}{@{}r@{}}$r_t = \frac{1}{h_{t+1}^{(1)}} h_{t+1}^{(0)}$ \\ $ -  \frac{1}{h_{t}^{(1)}} h_{t}^{(0)}$ \end{tabular}\\
  \bottomrule
\end{tabular}
\caption{Examples of non-cumulative objective functions $f$.}
\label{tab:ftable}
\end{table}
\setlength{\tabcolsep}{6pt}

\begin{wrapfigure}{R}{0.5\textwidth}
 \begin{center}
    \includegraphics[width=0.48\textwidth]{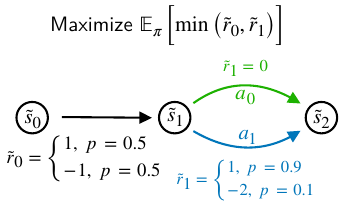}
    \caption{Two-step decision process with non-cumulative objective. In the first step, the agent has only one action available and receives a probabilistic reward. In the second step, the agent can choose between two actions.}\label{fig:fig2}
  \end{center}
  \vspace{-1\baselineskip}
\end{wrapfigure}

\begin{theorem}\label{theorem}
Consider an NCMDP $\tilde M$ and a corresponding MDP $M$, both defined as in \Cref{def:corr_MDP}. Then, given a policy $\pi(a_t | s_t)$ of $M$, the expected return of $M$ is equal to the expected return of $\tilde M$, i.e.\
\begin{equation}
    \mathbb{E}_\pi\left[\sum_{t=0}^{T-1}r_t\right] = \mathbb{E}_\pi\left[f(\tilde r_0,\dots, \tilde r_{T-1})\right],
\end{equation} if $\pi(a_t | s_t)$ is used to interact with $\tilde M$ as follows:
At each time step $t$, map the current trajectory $\mathcal{\tilde T}$ of $\tilde M$ to a trajectory $\mathcal{T}= \mathrm{map}\left(\mathcal{\tilde T}\right)$ of $M$ as described in \Cref{def:T_map}, thereby finding $s_t$. Then choose an action according to $\pi(a_t | s_t)$.
\end{theorem}
For a proof of this theorem see \Cref{app:proof}. Since $M$ is a standard MDP, we can find its optimal policy using standard methods and are guaranteed that it will also be the ideal policy of $\tilde M$.
More generally, also non-ideal policies will yield the same return for both decision processes.
By mapping the NCMDP to an MDP, all guarantees available for MDP solvers, e.g.\ convergence proofs for Q-learning \citep{watkins1992}, or the policy gradient theorem for policy-based methods \citep{sutton1999}, directly apply also to NCMDPs.

We demonstrate the need for the additional state information $h_t$ in the two-step NCMDP depicted in \Cref{fig:fig2}, which captures the main conceptual difficulties in solving NCMDPs: A non-cumulative objective in a stochastic environment.
Here, the goal is to maximize the expected minimum of the rewards, i.e. $\min(\tilde r_0, \tilde r_1)$. If $\tilde r_0 = 1$ in the first step, the ideal policy is to choose $a_1$ in the second step, while if $\tilde r_0 = -1$ the agent should choose $a_0$. Therefore, the choice of the ideal action depends on information contained in the past rewards, which is captured by $h_1 = \tilde r_0$. We show the MDP constructed from the NCMDP as described above in the Supplement in \Cref{fig:figA1} and its state-action value function found by value iteration in \Cref{tab:Qtable}. Our method finds the optimal policy, while previous methods for solving NCMDPs fail even in this simple example because they neglect the extra state information (see \Cref{tab:QtableCui}).

The question remains how to find the functions $u$ and $\rho$ given a new objective $f$. 
There is always at least one MDP corresponding to a given NCMDP since we can take $h_{t} = [\tilde r_0,\dots,\tilde r_{t-1}]$ with $u(h_t, \tilde r_t) = [h_t, \tilde r_t]$ and $\rho(h_{t}, \tilde r_t) = f([h_t, \tilde r_t]) - f(h_t)$. However, this is not ideal because it leads to a state size that grows linearly with the trajectory length. In \Cref{app:nec_cond}, we provide a necessary and sufficient condition for objectives $f$ with constant size additional state information $h_t$. Moreover, we also provide a sufficient, but not necessary, condition that is easier to verify and covers a large class of functions, e.g.\ all functions in \Cref{tab:ftable}. For functions in this class, we also provide an explicit construction of $u$, $\rho$, and $h_t$ (see \Cref{app:suff_cond}). In practice, we have empirically observed that a simple analytical consideration leads to $h_t$ of a small and constant size for each of the objectives $f$ considered in this manuscript, which is desirable for efficient learning and integration with standard function estimators such as neural networks. 

\textit{Limitation:} 
Our method may result in a significantly increased state space, which might pose challenges for tabular reinforcement learning methods and dynamic programming techniques. However, deep reinforcement learning techniques are well equipped to address these challenges, as they can effectively handle exponentially large and continuous state spaces.

From an implementation perspective, our scheme of mapping NCMDPs to MDPs requires minimal effort, since we can treat both the NCMDP and the used MDP solver as black boxes by simply adding a layer between them, as shown in \Cref{fig:fig1}. In addition, our treatment facilitates online learning and does not require any computationally expensive preprocessing of the NCMDP.
This opens the door for researchers with specific domain knowledge, who are not necessarily experts in reinforcement learning, to use standard libraries such as \texttt{stable-baselines3} \citep{stable-baselines3} to solve their non-cumulative problems. Conversely, it allows reinforcement learning experts to quickly tackle existing environments using our method.

\section{Experiments}\label{sec:experiments}

\begin{wrapfigure}{R}{0.5\textwidth}
 \begin{center}
    \includegraphics[width=0.48\textwidth]{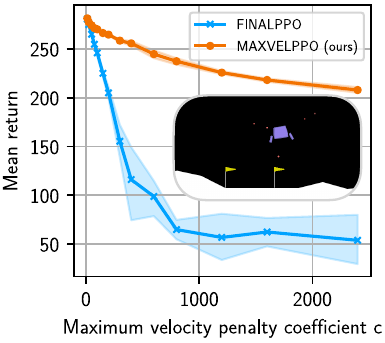}
    \caption{Average return in the lunar lander environment (inset) against the velocity penalty coefficient $c$ defined in \Cref{eq:ll_return} for the non-cumulative MAXVELPPO and the cumulative FINALPPO algorithm.}\label{fig:fig3}
  \end{center}
  \vspace{-2\baselineskip}
\end{wrapfigure}

There are three kinds of experiments presented in this manuscript:
\begin{enumerate}
    \item We show that our method, in conjunction with dynamic programming, can find the optimal policy in a proof-of-principle stochastic NCMDPs where previous methods fail (see previous section, \Cref{fig:fig2}).
    \item In the supplemental \Cref{app:cui_comp}, we compare our method with the only previous approach for solving NCMDPs with objectives other than the $\max$ function, which was introduced by \citet{cui2023}, and is only applicable with more restrictive objectives $f$, in deterministic environments, and in conjunction with Q-learning methods. Our experiments indicate that even in this restricted setting, our more general method may outperform the previous method in terms of final performance.
    \item In the following sections, we show applications in real-world tasks where previous methods used for solving special cases of NCMDPs cannot be applied. In lack of a general method for solving NCMDPs, state-of-the-art approaches to these real-world problems have so far relied on standard MDPs and approximate solutions which we use as a baseline to compare our method to. 
\end{enumerate}

If not otherwise specified, shaded regions in plots and uncertainties in tables denote a $95\%$ bootstrap confidence interval calculated over multiple seeds of the experiment. We provide details on hyperparameters, compute resources, and training curves for all experiments in the supplemental \Cref{app:details_experiments}. 

\subsection{Classical Control}\label{sec:lunar_lander}

As a first use case of our method, we train a reinforcement learning agent in the lunar lander environment of the \texttt{gymnasium} \citep{towers_gymnasium_2023} library. The agent controls a spacecraft with four discrete actions corresponding to different engines while being pushed by a stochastic wind. Immediate positive rewards $r_t$ are given for landing the spacecraft safely with small negative rewards given for using the engines. A realistic goal when landing a spacecraft is to not let the spacecraft get too fast, e.g.\ to avoid excessive frictional heating. Therefore, we define an objective where the agent is penalized for its maximum speed during a trajectory, i.e.\ we try to maximize 
\begin{equation}\label{eq:ll_return}
    \mathbb{E}_\pi \left[\sum_{t=0}^{T-1} r_t - c \max{(v_0, \dots, v_{T-1})}\right],
\end{equation}
where $v_t$ is the speed of the agent at time $t$ and $c$ defines a trade-off between minimizing the maximum speed and the other goals of the agent.
We train RL agents using Proximal Policy Optimization (PPO) \citep{schulman2017} on the MDP constructed from the NCMDP as described above (MAXVELPPO) for different values of $c$.
We compare our results to an RL agent where the velocity penalty $- c \max{(v_0, \dots, v_{T-1})}$ is added to the final reward resulting in a cumulative objective (FINALPPO).
As shown in \Cref{fig:fig3}, the non-cumulative MAXVELPPO agent can find a better trade-off than the cumulative FINALPPO agent.

The reward function of the FINALPPO agent is non-Markovian. Therefore, we also tested adding the extra state information $h_t$, i.e.\ the past maximum velocity, to the FINALPPO agent's observations to restore Markovianity. However, this did not result in a performance gain (see \Cref{fig:fig_train_ll}). 
All experiments were performed with 5 agents using different seeds. 

Our method presented here could be applied to similar use cases in other classical control problems, such as teaching a robot to reach a goal while minimizing the maximum impact forces on its joints or the forces its motors need to apply.

\subsection{Portfolio Optimization with Sharpe Ratio as Objective}

\begin{wrapfigure}{R}{0.5\textwidth}
 \begin{center}
    \includegraphics[width=0.48\textwidth]{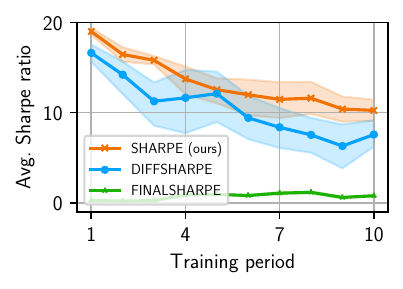}
    \caption{Portfolio optimization. Sharpe ratio during 10 different training periods for algorithms maximizing either the cumulative differential Sharpe ratio (DIFFSHARPE) or the exact Sharpe ratio by using our non-cumulative method (SHARPE) or by giving the Sharpe ratio as a single final reward (FINALSHARPE).}\label{fig:fig6}
  \end{center}
  \vspace{-1\baselineskip}
\end{wrapfigure}

Next, we consider the task of portfolio optimization where an agent decides how to best invest its assets across different possibilities. A common measure for a successful investment strategy is its Sharpe ratio \citep{sharpe1966}
\begin{equation*}
    \frac{\mathrm{MEAN}(\tilde r_0,\dots,\tilde r_{T-1})}{\mathrm{STD}(\tilde r_0,\dots,\tilde r_{T-1})},
\end{equation*}
where $\tilde r_t = (P_{t+1} - P_t) / P_t$ are the simple returns and $P_t$ is the portfolio value at time $t$. By dividing through the standard deviation of the simple returns, the agent is discouraged from risky strategies with high volatility. As the Sharpe ratio is non-cumulative, reinforcement learning strategies so far needed to fall back on the approximate differential Sharpe ratio as a reward \citep{moody1998, moody2001}. However, using the methods developed in this paper we can directly maximize the exact Sharpe ratio. 
We perform experiments on an environment as described by \citet{sood2023} where an agent trades on the 11 different S\&P500 sector indices between 2006 and 2021. The states contain a history of returns of each index and different volatility measures. Actions are the continuous relative portfolio allocations for each day.
The experiment described by \citet{sood2023} consists of training 5 agents with different seeds on 10 overlapping periods spanning 5 years each. We re-implement this experiment by training agents using PPO with the cumulative differential Sharpe ratio (DIFFSHARPE) or the exact Sharpe ratio as their objective. For the agents maximizing the exact Sharpe ratio, we either phrase the problem as an NCMDP and use our method described above (SHARPE), or we provide a single reward at the end of the trajectory (FINALSHARPE).
As depicted in \Cref{fig:fig6}, the SHARPE algorithm significantly outperforms the other algorithms. As in \Cref{sec:lunar_lander}, restoring Markovianity in the FINALSHARPE algorithm by adding $h_t$ to the states did not improve performance.

\textit{Limitation:} 
After training, all of the considered algorithms do not generalize well to new, unseen periods indicating over-fitting of the agents' policies to the years they are trained on. Therefore, the experiments presented here should be understood as showing improved performance of the SHARPE agent only in the environment it was trained in.
This limitation could potentially be offset by generating more data to train on, e.g.\ by using realistic stock-market simulators which are for example being developed using Generative Adversarial Networks \citep{Li_2020}.

While the Sharpe ratio is most widely adopted in finance, our method opens up the possibility to maximize it also in other scenarios where risk-adjusted rewards are desirable, i.e.\ all problems where consistent rewards with low variance are more important than a higher cumulative reward. For example, in chronic disease management, maintaining stable health metrics is preferable to sporadic improvements. In emergency or customer service, ensuring predictable response times is often more important than occasional fast responses mixed with slow ones.

\subsection{Discrete Optimization Problems}\label{sec:discrete_opt}

Next, we consider a large class of applications where RL is commonly used: Problems where the agent iteratively transforms a state by its actions to find a state with a lower associated cost. These problems are common in scientific applications such as physics, e.g.\ to reduce the length of quantum logic circuits \citep{fosel2021}, or chemistry, e.g.\ for molecular discovery \citep{zhou2019}. Another prominent example is the discovery of new algorithms \citep{Mankowitz2023}. Intuitively, these problems can be understood as searching for the state with the lowest cost within an equivalence class defined by all states that can be reached from the start state by the agent's actions.

Concretely, we consider the class of discrete optimization problems equipped with a scalar cost function $c(\tilde s_t)$ and the immediate rewards $\tilde r_t = c(\tilde s_t) - c(\tilde s_{t+1})$. Additionally, we are interested in the state with the lowest cost found during a trajectory, i.e.\ the goal is to maximize
\begin{equation}\label{eq:max_sum}
    \mathbb{E}_\pi \left[c(\tilde s_0) -  \underset{k\in[0, T-1]}{\min} c(\tilde s_k) \right]= \mathbb{E}_\pi \left[\underset{k\in[-1, T-1]}{\max}\sum_{t=0}^k \tilde r_t \right].
\end{equation}
In the deterministic settings we study in the following experiments, the ideal policy for maximizing the cumulative reward $\mathbb{E}_\pi \left[\sum_{t=0}^{T-1} \tilde r_t \right]$ 
maximizes also \Cref{eq:max_sum}.
We conjecture that directly maximizing \Cref{eq:max_sum} by formulating the problem as an NCMDP will yield better results due to the following reasons:

\begin{enumerate}
    \item The agent does not need to learn an optimal stopping point.
    \item Considering only the rewards up to the minimum found cost might decrease the variance of the gradient estimate.
    \item The agent does not receive negative rewards for trying to escape a local cost minimum during a trajectory and is therefore not discouraged from exploring. This leads to learning difficult optimization strategies requiring an intermittent cost increase more easily.
\end{enumerate}

\begin{figure}[t]
    \centering
    \includegraphics{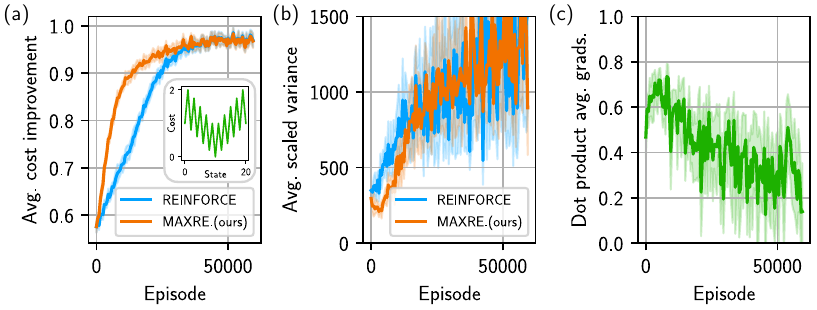}
    \caption{Peak environment. (a) Cost improvement during the trajectory against training episodes for the cumulative REINFORCE agent and the MAXREINFORCE agent maximizing \Cref{eq:max_sum}. The inset shows the cost function of the environment. (b) Empirical variance of both agents against training progress. (c) Estimated dot product of the average gradient of both agents against training progress.}
    \label{fig:fig4}
\end{figure}

\subsubsection{Peak environment}

To facilitate an in-depth analysis, we first consider a toy environment with the cost function depicted in the inset of \Cref{fig:fig4}\,(a). The cost function was chosen to be simple while still requiring intermittently cost-increasing actions of the optimal policy. Each trajectory lasts 10 steps and the agent's actions are stepping to the left, right, or doing nothing. To minimize the number of hyperparameters, we use the REINFORCE algorithm \citep{williams1992} with a tabular policy.
We compare agents trained with the non-cumulative objective \Cref{eq:max_sum} (MAXREINFORCE) with agents that maximize the cumulative rewards (REINFORCE). As shown in \Cref{fig:fig4}\,(a), the MAXREINFORCE agent trains much faster. Two possible sources of this speed-up are a different, more advantageous direction of the gradient updates and a reduced variance when estimating these gradients. To investigate which is the case, we periodically stop training and run $n=1000$ trajectories with a fixed policy. We then compute the empirical variance of the gradient update inspired by \citet{kaledin2022} as 
\begin{equation*}
    \mathrm{VAR} = \frac{1}{n} \sum_{i=0}^{n-1} \norm{\vec g_i}_2^2 - \norm{ \frac{1}{n} \sum_{i=0}^{n-1} \vec g_i}_2^2,
\end{equation*}
where $\vec g_i$ is the gradient from a single trajectory derived either by the MAXREINFORCE or the REINFORCE algorithm. As this variance scales with the squared magnitude of the average gradients, we normalize it by this value to ensure a fair comparison between the two algorithms. In \Cref{fig:fig4}\,(b), we show that in the initial phase of training, the MAXREINFORCE algorithm significantly reduces variance. To compare the direction of the gradients we use the same data to compute the normalized average gradients of both algorithms and show their dot product in  \Cref{fig:fig4}\,(c). We find that the gradients are correlated (i.e.\ the dot product is bigger than zero) but not the same. Therefore, we conclude that the training speed-up is derived both from lower variance and a better true gradient direction.
All results reported are averaged over 10 seeds.

\subsubsection{Quantum error correction}
As a first real-world use case, we focus on optimization problems where the initial state is always the same and the optimal stopping point is known. The specific task we consider is the search for quantum programs to prepare logical states of a given quantum error correction code - critical for the eventual realization of quantum computation \citep{Terhal2015}. The agent iteratively adds elementary quantum logic gates until the program delivers the desired result. The agents' policies are encoded by standard multilayer perceptrons and they are trained on five different sized problems. For details on the used environment, see \citet{zen2024}.
We use PPO to train agents to maximize either \Cref{eq:max_sum} (MAXPPO) or the cumulative reward (PPO).
To obtain a single performance measure encompassing both final performance and training speed, we continuously evaluate the mean cost improvement of the agents and average it over the training process as suggested by \citet{andrychowicz2020}. Intuitively, this measure can be understood as the area under the agent's training curve. 
In \Cref{tab:qectable}, we report the quotient of this performance measure of the MAXPPO algorithm and the PPO algorithm for both the best of 10 and the mean of 10 trained agents.
The different tasks are denoted by three integers, as customary in the quantum error correction community. We find that the MAXPPO algorithm performs significantly better.

We also trained MAXPPO for a second quantum circuit discovery task [for details see \citet{olle2023}]. In this case, PPO and MAXPPO performed equally well. We argue that the large degeneracy present in the solution space of this second discovery task diverts the more exploratory MAXPPO, lowering its performance to the level of PPO.

\begin{table}[t]
\centering
\begin{tabular}{ccc} 
\toprule
\multirow{2}[3]{*}{Task} & \multicolumn{2}{c}{Area under training curve $\frac{\mathrm{MAXPPO}}{\mathrm{PPO}}$}\\
\cmidrule(lr){2-3}
 & Best of 10 & Mean of 10 \\ 
\midrule
$[[5,1,3]]$ & $1.04$ & $1.07\,(1.04, 1.09)$\\
$[[7,1,3]]$ & $1.18$ & $1.21\,(1.17, 1.25)$\\
$[[9,1,3]]$ & $1.01$ & $1.09\,(1.04, 1.17)$\\
$[[15,1,3]]$ & $1.14$ & $1.03\,(0.98, 1.08)$\\
$[[17,1,5]]$ & $1.47$ & $1.62\,(1.52, 1.78)$\\
\bottomrule
\end{tabular}
\caption{Results for quantum error correction environments.}\label{tab:qectable}
\end{table}

\subsubsection{ZX-diagrams}\label{zx-diagrams}

\begin{figure}[t]
    \centering
    \includegraphics{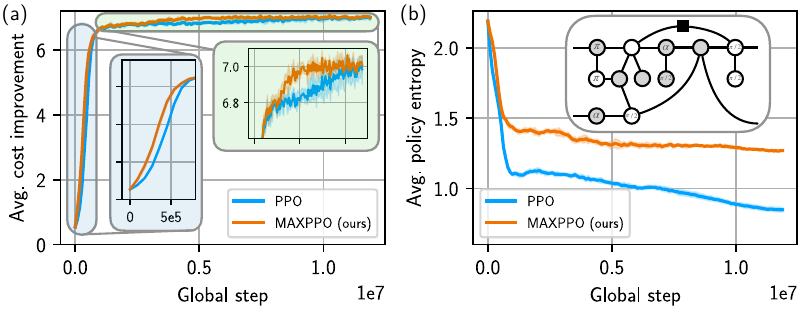}
    \caption{ZX environment. (a) Cost improvement during the trajectory against total steps taken in the environment for the cumulative PPO agent and the MAXPPO agent maximizing \Cref{eq:max_sum}. Left inset: Zoom-in to the initial training phase. Right inset: Zoom-in to the final training phase. (b) Entropy of the agents' policies against total steps taken in the environment. Inset: A typical ZX-diagram.}
    \label{fig:fig5}
\end{figure}

As another real-world discrete optimization problem, we consider the simplification of ZX-diagrams, which are graph representations of quantum processes \citep{Coecke_2017} with applications e.g.\ in the compilation of quantum programs \citep{duncan2020, riu2023}. An example of a typical ZX-diagram is shown in the inset of \Cref{fig:fig5}\,(b). We consider the environment described by \citet{nagele2023}, where
the cost function of a diagram is given by its node number, the start states are randomly sampled ZX-diagrams, and the actions are a set of local graph transformations. In total, there are 6 actions per node and 6 actions per edge in the diagram. This is a challenging reinforcement learning task that requires the use of graph neural networks to accommodate the changing size of the state and action space. As shown in \Cref{fig:fig5}\,(a) the MAXPPO agent initially trains faster than the PPO agent (left inset). This is likely due to the reduced variance and different gradient direction as described above. The PPO agent then shortly catches up, but ultimately requires about twice as many training steps to reach optimal performance as the MAXPPO agent (right inset). We argue that this is because the MAXPPO agent is better at exploring and therefore learning difficult optimization strategies (reason 3 in \Cref{sec:discrete_opt}). This is captured by the entropy of the MAXPPO agent's policy staying much higher than the entropy of the PPO agent, as shown in \Cref{fig:fig5}\,(b). The reported results are averaged over 5 seeds.

\textit{Limitation:} When applying MAXPPO to discrete optimization problems with long trajectories in the many hundreds of steps, we empirically observed an initially slow learning speed. This could be due to the agent initially mostly increasing cost and therefore receiving zero reward for almost the entire trajectory. A possible solution could be to dynamically adjust the trajectory length in these problems going from shorter to longer trajectories during the training process.

\section{Related Work}\label{sec:related_work}

A special case of NCMDPs is first considered by \citet{quah2006}, who adapt Q-learning to the $\max$ objective by redefining the temporal difference error of their learning algorithms and demonstrate their algorithms on an optimal stopping problem. However, they do not adapt state space and do not provide theoretical convergence guarantees.  \citet{gottipati2020} rediscover the same algorithm, apply it to molecule generation, and provide convergence guarantees for their method, while \citet{eyckerman2022} consider the same algorithm for application placement in fog environments. \citet{cui2023} finally show a shortcoming of the method used in the above papers: It is guaranteed to converge to the optimal policy only in deterministic settings. Additionally, they provide convergence guarantees of Q-learning in deterministic environments for a larger class of functions $f$, focusing on the $\min$ function for network routing applications. In \Cref{app:nec_cond} we show that the class of objectives $f$ considered by \citet{cui2023} is a subclass of all objectives that lead to constant size extra state information $h_t$ using our method.
Independently, the $\max$ function is also used in the field of safety reinforcement learning in deterministic environments both for Q-learning \citep{fisac2015, fisac2019, hsu2021} and policy-based reinforcement learning \citep{yu2022}.
\citet{moflic2023} use a reward function with parameters that depend on the past rewards of the trajectory to tackle quantum circuit optimization problems that require large intermittent negative rewards, albeit without providing theoretical convergence guarantees.
Additionally, \citet{Wang2020} investigate planning problems with non-cumulative objectives within deterministic settings. They provide provably efficient planning algorithms for a large class of functions $f$ by discretizing rewards and appending them to the states.
Recently, \citet{veviurko2024} showed how to use the $\max$ objective also in probabilistic settings by augmenting state space and provide convergence guarantees for both Q-learning and policy-based methods. They then show experiments with their algorithm yielding improvements in MDPs with shaped rewards. However, they do not redefine the rewards, which requires adaptation of the implementation of their MDP solvers.  Our method described above reduces to an effectively equivalent algorithm to theirs in the special case of the $\max$ objective. 

A limitation of all works discussed above is that they require a potentially complicated adaptation of their reinforcement learning algorithms and only consider specific MDP solvers or specific non-cumulative objectives. They are also limited to deterministic settings, except for \citet{veviurko2024}, which consider only the $\max$ objective.

\section{Discussion \& Conclusion}

In this work, we described a mapping from a decision process with a general non-cumulative objective (NCMDP) to a standard Markov decision process (MDP) applicable in deterministic and probabilistic settings. As our method is agnostic to the algorithm used to solve the resulting MDP, it works for arbitrary action spaces and in conjunction with both off- and on-policy algorithms.
Its implementation is straightforward and directly enables solving NCMDPs with state-of-the-art MDP solvers, allowing us to show improvements in a diverse set of tasks such as classical control problems, portfolio optimization, and discrete optimization problems. Note that these improvements are achieved without adding a single additional hyperparameter to the solving algorithms. Moreover, the computational overhead of our method is typically negligible, since it requires only the evaluation of a few elementary mathematical expressions per time step in the environment (see \Cref{tab:ftable}).

In further theoretical work, a full constructive classification of objective functions $f$ with constant-size extra state information $h_t$ would be desirable. From an applications perspective, there are a lot of interesting objectives with non-cumulative $f$ that could not be maximized so far. For example, the geometric mean could be used to maximize average growth rates, or the function $ f(\tilde r_0, \dots, \tilde r_{t}) = \delta^{t} \sum_{k=0}^{t} \tilde r_k, \delta \in (0,1)$ could be used to define an exponential trade-off between trajectory length and cumulative reward in settings where long trajectories are undesirable. We believe that a multitude of  other applications with non-cumulative objectives are still unknown to the reinforcement learning  (RL) community, and conversely, that researchers working on non-cumulative problems are not aware of RL, simply because these two concepts could not straightforwardly be unified so far. This manuscript offers the exciting possibility of discovering and addressing this class of problems still unexplored by reinforcement learning.

\section{Code Availability}

We open-source all code used to produce the results of this manuscript in the following GitHub repositories: Classical control, portfolio optimization, peak environment, and grid environment (\Cref{app:cui_comp}): \cite{githubncmdp}. ZX-diagrams: \cite{githubzx}. Quantum error correction code discovery: \cite{githubqec}. Logical state preparation: \cite{githubls}.

\section*{Acknowledgments}
This research is part of the Munich Quantum Valley (K-4 and K-8), which is supported by the Bavarian state government with funds from the Hightech Agenda Bayern Plus.
\begin{appendices}
\appendixpage
\renewcommand\thefigure{A\arabic{figure}}    
\setcounter{figure}{0}  
\renewcommand\thetable{A\arabic{table}}    
\setcounter{table}{0}

\section{Proof of Theorem 1}\label{app:proof}
First, we find that the probability $p(\mathcal{T})$ of a trajectory $\mathcal{T}$ occurring in $M$ is the sum of the probabilities $p(\mathcal{\tilde T})$ of all possible trajectories $\mathcal{\tilde T}$ occurring in $\tilde M$ that map to $\mathcal{T}$ if $\pi$ is used as a policy for $\tilde M$ as described above:
\begin{equation}\label{eq:p_t}
    \begin{split}
        p(\mathcal{T}) &= p(s_0) \prod_{t=0}^{T-1} \pi(a_t|s_t) p(r_t, s_{t+1}| s_t, a_t)
        \\
        &= p(\tilde s_0) \prod_{t=0}^{T-1} \pi(a_t|s_t) 
        \sum_{\tilde r_t} \tilde p(\tilde r_t, \tilde s_{t+1}| \tilde s_t, a_t) \delta_{h_{t+1},u(h_{t}, \tilde r_t)} \delta_{r_t,\rho(h_{t}, \tilde r_t)}
        \\
        &= \sum_{\mathrm{map}\left(\mathcal{\tilde T}\right) = \mathcal{T}}
        p(\tilde s_0) \prod_{t=0}^{T-1} \pi(a_t|s_t) \tilde p(\tilde r_t, \tilde s_{t+1}| \tilde s_t, a_t)
        = \sum_{\mathrm{map}\left(\mathcal{\tilde T}\right) = \mathcal{T}} p(\tilde T),
    \end{split}
\end{equation}
where $\sum_{\mathrm{map}\left(\mathcal{\tilde T}\right) = \mathcal{T}}$ is the sum over all trajectories of $\tilde M$ that map to $\mathcal{T}$. The second to last step can be shown through induction.
Therefore, we find
\begin{equation*}
\begin{split}
    \mathbb{E}_\pi\left[\sum_{t=0}^{T-1}r_t\right] &= \sum_{\mathcal{T}} p(\mathcal{T}) \sum_{t=0}^{T-1} r_t = \sum_\mathcal{T} \sum_{\mathrm{map}\left(\mathcal{\tilde T}\right) = \mathcal{T}} p(\mathcal{\tilde T}) f(\tilde r_0, \dots, \tilde r_{T-1})\\
    &= \sum_{\mathcal{\tilde T}} p(\mathcal{\tilde T}) f(\tilde r_0, \dots, \tilde r_{T-1}) = \mathbb{E}_\pi\left[f(\tilde r_0,\dots, \tilde r_{T-1})\right],
\end{split}
\end{equation*}
where in the second step we used \Cref{eq:r_def,eq:p_t}, and in the third step that each $\mathcal{\tilde T}$ maps to a unique $\mathcal{T}$. \hfill $\blacksquare$

\section{Objectives $f$ with Constant Size Extra State Information $h_t$}\label{app:cons_size_h}

\subsection{Necessary and sufficient condition for constant size extra state information~$h_t$}\label{app:nec_cond}
First, note that if we can predict $f(\tilde r_0, \dots, \tilde r_t)$ in a Markovian manner, we can also predict $f(\tilde r_0, \dots, \tilde r_t) - f(\tilde r_0, \dots, \tilde r_{t-1})$ in a Markovian manner by adding just one more dimension to our state $h_{t}$, i.e. $f(\tilde r_0, \dots, \tilde r_{t-1})$. Therefore, we focus on predicting $f(\tilde r_0, \dots, \tilde r_t)$ in the following. Rewriting \Cref{def:corr_MDP}, we find using
\begin{equation*}\label{eq:def_unrolled}
	f(\tilde r_0, \dots, \tilde r_t) - f(\tilde r_0, \dots, \tilde r_{t-1}) = \rho\left(\tilde r_t, u\left(\tilde r_{t-1}, u\left(\tilde r_{t-2}, ...\right)\right)\right)
\end{equation*}
or equivalently
\begin{equation}\label{eq:comptocui}
f(\tilde r_0, \dots, \tilde r_t)  = \rho^\prime\left(\tilde r_t, u^\prime\left(\tilde r_{t-1}, u^\prime\left(\tilde r_{t-2}, ...\right)\right)\right),
\end{equation}
with
\begin{equation*}
u^\prime(\tilde r_t, h_t^\prime) = u^\prime \left(\tilde r_t, (h_t, f_{t-1}) \right) =  \left[u(\tilde r_t, h_t), f_{t-1} + \rho(\tilde r_{t}, h_{t})\right],
\end{equation*}
and
\begin{equation*}
 \rho^\prime(\tilde r_t, h^\prime_t) = \rho(\tilde r_t, h_t) + f_{t-1}.
\end{equation*}
Given these definitions $f_{t} = f(\tilde r_0, \dots, \tilde r_t)$.
If $f$ admits a representation of the form \Cref{eq:comptocui} with $u^\prime$ of constant output dimension, additional state information $h_t$ of the same constant size is possible. \citet{cui2023} find a condition of similar form to \Cref{eq:comptocui} for the objectives $f$ which their method can optimize:
\begin{equation*}
 f(\tilde r_0, \dots, \tilde r_t) = g\left(\tilde r_t, g\left(\tilde r_{t-1}, g\left(\tilde r_{t-2}, \dots\right)\right)\right), g: \mathbb{R}^2 \rightarrow \mathbb{R}.
\end{equation*} 
In this sense, our method can be seen as extending the class of objectives $f$ with constant size extra state information by allowing for $\rho \neq u$ and by allowing a multidimensional update function $u: \mathbb{R}^{k+1} \rightarrow \mathbb{R}^k$ instead of $g$. Also, note that the method of \citet{cui2023} only works in deterministic environments in conjunction with Q-learning based methods.

\subsection{Sufficient condition for constant size extra state information $h_t$}\label{app:suff_cond}

While \Cref{eq:comptocui} provides a complete categorization of functions with constant size $h_t$, in practice it may be difficult to check whether a given function satisfies this property. In the following, we consider a smaller set of functions including all of the functions in \Cref{tab:ftable}, and provide an explicit construction of $u$, constant size $h_t$, and $\rho$ for functions of this class. Specifically, we consider function families that can be written with a constant $k\in\mathbb{N}$ as
\begin{equation}\label{eq:f_subclass}
f(\tilde r_0, \dots, \tilde r_t) = F\left(t, b_0, \dots, b_{k-1}\right),
\end{equation}
where $F: \mathbb{R}^{k+1} \rightarrow \mathbb{R}$,
\begin{equation*}
    b_j = b_j(\tilde r_0, \dots, \tilde r_t) = \mathbb{B}_j\big(\varphi_j(0, \tilde r_0), \dots, \varphi_j(t, \tilde r_t)\big),
\end{equation*}
where $\varphi_j : \mathbb{R}^2 \rightarrow \mathbb{R}$, and $\mathbb{B}_j$ is an arbitrary binary operation, such as $+, \times, \max, \min$, and 
\begin{equation*}
   \mathbb{B}_j(x_0, \dots, x_t) = \mathbb{B}_j(x_t, \mathbb{B}_j(x_{t-1}, \mathbb{B}_j(...))).
\end{equation*}
For example, if $\mathbb{B}_j$ is the multiplication operation, $\mathbb{B}_j(x_0, \dots, x_t) = \prod_{i=0}^{t} x_i$. As we show by construction below, all objectives $f$ of this form have extra state information $h_t$ of maximum dimension $k + 1$. Since we are dealing with function families, \Cref{eq:f_subclass} introduces a notion of consistency for different sizes of inputs $(\tilde r_0, \dots, \tilde r_t)$, since it requires the use of the same $\varphi_j$ for all input sizes. Note that for permutation invariant $f$, $\varphi_j$ is independent of the time step, i.e.\ $\varphi_j(i, \tilde r_i) =\varphi_j(\tilde r_i)$, and $\mathbb{B}_j$ is associative.
All but one of the objectives $f$ we consider in this manuscript are permutation invariant.

Given \Cref{eq:f_subclass}, we can directly construct the update function 
\begin{equation*}
h_{t+1}^{(j)} = u^{(j)}\left(\tilde r_t, h_t \right) =\mathbb{B}_j\left(\varphi_j\left(h_{t}^{(k)}, \tilde r_t\right), h_t^{(j)}\right), 0 \leq j < k,
\end{equation*}
where superscript $(j)$ indicates the $j$th entry of a vector. 
Additionally,
$$h_{t+1}^{(k)} = h_{t}^{(k)} + 1$$
is keeping track of the time step.
In this case, each $b_j(\tilde r_0, \dots, \tilde r_{t-1})$ is a dimension of $h_t$. Finally,
\begin{equation*}
\rho(\tilde r_t, h_t) = F\left(h_{t+1}^{(k)}, \left\{h_{t+1}^{(j)}\right\}_{j=0}^{j=k-1}\right) - F\left(h_{t}^{(k)}, \left\{h_{t}^{(j)}\right\}_{j=0}^{j=k-1}\right).
\end{equation*}
We show in \Cref{tab:fansatze} how all functions in \Cref{tab:ftable} can be written in this form.

\begin{table}[t]
\centering
\begin{tabular}{cccc}
 \toprule
 $f(\tilde r_0, \dots , \tilde r_{t})$ & $F\left(t, b_0, \dots, b_{k-1}\right)$ & $[\mathbb{B}_j]$ & $[\varphi_j(t, \tilde r_t)]$ \\ 
  \toprule
  $\max(\tilde r_0,\dots,\tilde r_{t})$ & $h_t^{(0)}$ &  $\mathbb{B}_0 = \max$ & $\varphi_0(\tilde r_t) = \tilde r_t$   \\  
   \midrule
  $\min(\tilde r_0,\dots,\tilde r_{t})$ & $h_t^{(0)}$ & $\mathbb{B}_0 = \min$ & $\varphi_0(\tilde r_t) = \tilde r_t$  \\  
  \midrule
 \makecell[c]{Sharpe ratio\\  $\frac{\mathrm{MEAN}(\tilde r_0,\dots,\tilde r_{t})}{\mathrm{STD}(\tilde r_0,\dots,\tilde r_{t})}$} & $\frac{h_t^{(0)}/t}{\sqrt{h_t^{(1)}/t-\left(h_t^{(0)}/t\right)^2}}$ & $\mathbb{B}_0 = \mathbb{B}_1 = + $ & \makecell[c]{$\varphi_0(\tilde r_t) = \tilde r_t$, \\$\varphi_1(\tilde r_t) = \tilde r_t^2$}\\
 \midrule
  $\underset{k\in[-1, t]}{\max}\sum_{i=0}^k \tilde r_i$ & $h_t^{(1)} + h_t^{(0)}$  & \makecell[l]{$\mathbb{B}_0\left(\varphi_0(\tilde r_t), h_t^{(0)}\right) =$ \\$  \max\left(0, h_t^{(0)} - \varphi_0(\tilde r_t)\right),$ \\ $\mathbb{B}_1= + $} &  \makecell[c]{$\varphi_0(\tilde r_t) = \tilde r_t$, \\$\varphi_1(\tilde r_t) = \tilde r_t$} \\ 
 \midrule
 $\tilde r_0 \tilde r_1 \dots \tilde r_{t}$ & $h_t^{(0)}$ &$\mathbb{B}_0 = \times $ & $\varphi_0(\tilde r_t) = \tilde r_t$\\ 
 \midrule
 \makecell[c]{Harmonic mean \\ $\frac{1}{\frac{1}{\tilde r_0} + \dots + \frac{1}{\tilde r_{t}}}$} & $\frac{1}{h_t^{(0)}}$ & $\mathbb{B}_0 = +$ & $\varphi_0(\tilde r_i) = \frac{1}{\tilde r_t}$ \\ 
 \midrule
 \begin{tabular}{@{}c@{}} $\delta^{t} \sum_{t=0}^{t} \tilde r_t$, \\ $ \delta \in (0,1)$\end{tabular} & $\delta^{t} h_t^{(0)}$ & $\mathbb{B}_0 = +$ & $\varphi_0(\tilde r_t) = \tilde r_t$\\
  \midrule
  $\frac{1}{t+1} \sum_{k=0}^{t} \tilde r_k$ & $\frac{1}{t} h_t^{(0)}$ & $\mathbb{B}_0 = +$ & $\varphi_0(\tilde r_t) = \tilde r_t$\\
  \bottomrule
\end{tabular}
\caption{Objectives $f$ in the form of \Cref{eq:f_subclass}. All $f$ except $\underset{k\in[-1, t]}{\max}\sum_{i=0}^k \tilde r_i$ are permutation invariant resulting in associative $\mathbb{B}_j$.}
\label{tab:fansatze}
\end{table}

\section{Grid Environment: Experiments Comparing to \citet{cui2023}}\label{app:cui_comp}

\begin{figure}[t]
    \centering
    \includegraphics{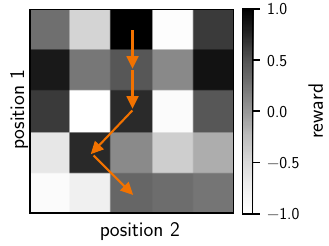}
    \caption{Example of a grid environment for the $\min$ objective with $N=5$. Orange arrows indicate the ideal policy.}
    \label{fig:reward_map_grid}
\end{figure}

\begin{table}[t]
\centering
\begin{tabular}{ccc}
 \toprule
 Grid seed & Final return, this work & Final return, Cui et\ al.\ \\
 \midrule
 \multicolumn{3}{c}{Grid size $N=3$} \\
 \midrule
0 & $\mathbf{0.46\,(0.46, 0.46)}$ & $0.16\,(0.09, 0.31)$ \\
1 & $0.66\,(0.66, 0.66)$ & $0.66\,(0.66, 0.66)$ \\
2 & $-0.45\,(-0.45, -0.45)$ & $-0.45\,(-0.45, -0.45)$ \\
3 & $-0.04\,(-0.04, -0.04)$ & $-0.06\,(-0.10, -0.04)$ \\
4 & $0.21\,(0.21, 0.21)$ & $0.21\,(0.21, 0.21)$ \\
5 & $-0.43\,(-0.43, -0.43)$ & $-0.43\,(-0.43, -0.43)$ \\
6 & $\mathbf{0.36\,(0.36, 0.36)}$ & $0.28\,(0.27, 0.32)$ \\
7 & $0.63\,(0.61, 0.64)$ & $0.63\,(0.61, 0.64)$ \\
8 & $-0.12\,(-0.12, -0.12)$ & $-0.12\,(-0.12, -0.12)$ \\
9 & $\mathbf{0.83\,(0.83, 0.83)}$ & $0.59\,(0.48, 0.72)$ \\
mean & $0.21$  & $0.15$\\
 \midrule
 \multicolumn{3}{c}{Grid size $N=4$} \\
 \midrule
0 & $\mathbf{0.44\,(0.21, 0.66)}$ & $0.09\,(0.09, 0.09)$ \\
1 & $0.13\,(-0.13, 0.39)$ & $-0.09\,(-0.09, -0.09)$ \\
2 & $\mathbf{0.31\,(0.31, 0.31)}$ & $0.20\,(0.12, 0.28)$ \\
3 & $-0.21\,(-0.44, -0.06)$ & $-0.08\,(-0.12, -0.04)$ \\
4 & $0.60\,(0.60, 0.60)$ & $0.60\,(0.59, 0.60)$ \\
5 & $-0.18\,(-0.18, -0.18)$ & $-0.18\,(-0.18, -0.18)$ \\
6 & $0.32\,(0.28, 0.35)$ & $0.33\,(0.30, 0.35)$ \\
7 & $-0.11\,(-0.11, -0.11)$ & $-0.10\,(-0.11, -0.08)$ \\
8 & $-0.43\,(-0.51, -0.28)$ & $-0.23\,(-0.36, -0.14)$ \\
9 & $-0.03\,(-0.03, -0.03)$ & $-0.05\,(-0.09, -0.03)$ \\
mean & $0.08$ & $0.05$ \\
 \midrule
 \multicolumn{3}{c}{Grid size $N=5$} \\
 \midrule
0 & $0.09\,(-0.03, 0.19)$ & $-0.08\,(-0.15, 0.07)$ \\
1 & $-0.09\,(-0.09, -0.09)$ & $-0.17\,(-0.32, -0.09)$ \\
2 & $-0.52\,(-0.59, -0.45)$ & $-0.45\,(-0.45, -0.45)$ \\
3 & $-0.00\,(-0.20, 0.23)$ & $-0.18\,(-0.31, -0.08)$ \\
4 & $\mathbf{0.60\,(0.60, 0.60)}$ & $0.06\,(-0.00, 0.20)$ \\
5 & $-0.18\,(-0.18, -0.18)$ & $-0.18\,(-0.18, -0.18)$ \\
6 & $\mathbf{0.36\,(0.36, 0.36)}$ & $-0.14\,(-0.14, -0.14)$ \\
7 & $-0.62\,(-0.81, -0.51)$ & $-0.35\,(-0.55, -0.13)$ \\
8 & $-0.52\,(-0.52, -0.52)$ & $\mathbf{-0.49\,(-0.49, -0.49)}$ \\
9 & $-0.03\,(-0.03, -0.03)$ & $-0.03\,(-0.03, -0.03)$ \\
mean & $-0.09$ & $-0.20$\\
  \bottomrule
\end{tabular}
\caption{Final return in the grid environment using our method or the method described by \citet{cui2023}. Statistically significant differences between the algorithms are highlighted.}
\label{tab:cui_vs_us_final_return}
\end{table}

\begin{table}[t]
\centering
\begin{tabular}{ccc}
 \toprule
 Grid seed &  \begin{tabular}{@{}c@{}}Area under training curve, \\ this work\end{tabular}  & \begin{tabular}{@{}c@{}}Area under training curve, \\ Cui et\ al.\ \end{tabular}  \\
 \midrule
 \multicolumn{3}{c}{Grid size $N=3$} \\
 \midrule
0 & $\mathbf{0.44\,(0.42, 0.45)}$ & $0.15\,(0.13, 0.17)$ \\
1 & $\mathbf{0.65\,(0.64, 0.65)}$ & $0.63\,(0.63, 0.63)$ \\
2 & $-0.46\,(-0.47, -0.46)$ & $\mathbf{-0.46\,(-0.47, -0.46)}$ \\
3 & $-0.06\,(-0.06, -0.06)$ & $-0.07\,(-0.11, -0.05)$ \\
4 & $0.18\,(0.16, 0.19)$ & $\mathbf{0.20\,(0.20, 0.21)}$ \\
5 & $-0.44\,(-0.44, -0.44)$ & $\mathbf{-0.44\,(-0.44, -0.44)}$ \\
6 & $0.34\,(0.33, 0.35)$ & $0.29\,(0.25, 0.33)$ \\
7 & $0.59\,(0.57, 0.60)$ & $\mathbf{0.57\,(0.56, 0.58)}$ \\
8 & $-0.13\,(-0.14, -0.13)$ & $\mathbf{-0.13\,(-0.14, -0.13)}$ \\
9 & $0.79\,(0.78, 0.81)$ & $0.55\,(0.53, 0.56)$ \\
mean & $0.19$ & $0.13$\\
 \midrule
 \multicolumn{3}{c}{Grid size $N=4$} \\
 \midrule
0 & $0.12\,(0.05, 0.21)$ & $0.05\,(0.02, 0.10)$ \\
1 & $-0.05\,(-0.15, 0.08)$ & $-0.15\,(-0.17, -0.13)$ \\
2 & $0.22\,(0.17, 0.27)$ & $0.23\,(0.20, 0.26)$ \\
3 & $-0.15\,(-0.16, -0.15)$ & $\mathbf{-0.12\,(-0.15, -0.09)}$ \\
4 & $\mathbf{0.56\,(0.54, 0.58)}$ & $0.48\,(0.45, 0.52)$ \\
5 & $-0.26\,(-0.28, -0.23)$ & $\mathbf{-0.21\,(-0.22, -0.20)}$ \\
6 & $0.13\,(0.05, 0.21)$ & $\mathbf{0.21\,(0.15, 0.26)}$ \\
7 & $-0.20\,(-0.24, -0.16)$ & $\mathbf{-0.14\,(-0.16, -0.13)}$ \\
8 & $-0.50\,(-0.51, -0.49)$ & $\mathbf{-0.34\,(-0.42, -0.28)}$ \\
9 & $-0.12\,(-0.14, -0.10)$ & $\mathbf{-0.06\,(-0.07, -0.05)}$ \\
mean & $-0.02$ & $0.0$ \\
 \midrule
 \multicolumn{3}{c}{Grid size $N=5$} \\
 \midrule
0 & $\mathbf{-0.03\,(-0.13, 0.07)}$ & $-0.16\,(-0.18, -0.14)$ \\
1 & $-0.11\,(-0.12, -0.10)$ & $-0.12\,(-0.12, -0.11)$ \\
2 & $-0.55\,(-0.59, -0.52)$ & $\mathbf{-0.48\,(-0.48, -0.47)}$ \\
3 & $-0.24\,(-0.32, -0.14)$ & $-0.21\,(-0.25, -0.16)$ \\
4 & $0.35\,(0.15, 0.50)$ & $0.20\,(0.07, 0.33)$ \\
5 & $-0.22\,(-0.22, -0.21)$ & $\mathbf{-0.20\,(-0.20, -0.20)}$ \\
6 & $-0.02\,(-0.14, 0.10)$ & $-0.14\,(-0.16, -0.10)$ \\
7 & $-0.69\,(-0.76, -0.61)$ & $\mathbf{-0.57\,(-0.65, -0.49)}$ \\
8 & $-0.53\,(-0.53, -0.53)$ & $\mathbf{-0.52\,(-0.52, -0.51)}$ \\
9 & $-0.13\,(-0.21, -0.08)$ & $\mathbf{-0.07\,(-0.08, -0.06)}$ \\
mean & $-0.09$ & $-0.20$ \\
  \bottomrule
\end{tabular}
\caption{Area under the training curve in the grid environment using our method or the method described by \citet{cui2023}. Statistically significant differences between the algorithms are highlighted.}
\label{tab:cui_vs_us_auc}
\end{table}

\begin{figure}[t]
    \centering
    \includegraphics[width=0.8\textwidth]{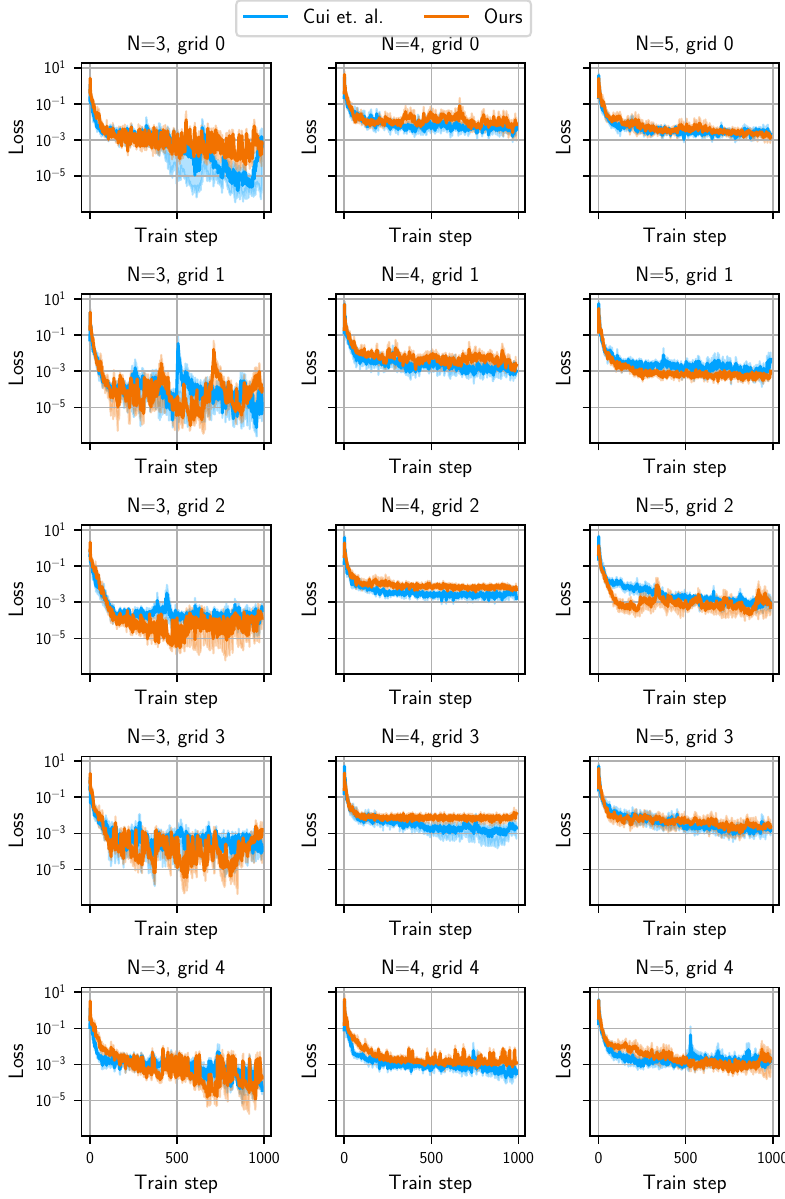}
    \caption{Square error when predicting the Q-function in the grid environment using our method (orange) and the method of \citet{cui2023} (blue) against the training step. $N$ indicates the grid size with 10 different random grids per size (only 5 shown). The prediction loss of the Q-function is similar for both methods but our method may lead to a higher final return (see \Cref{tab:cui_vs_us_final_return}).}
    \label{fig:losses_cui_comp}
\end{figure}

In this section, we compare our method to the more specialized method for solving NCMDPs introduced by \citet{cui2023}, which can only be applied in deterministic environments in conjunction with Q-learning methods. We want to answer the question of whether our more general method can be competitive even in this specialized scenario. To this end, we consider the $\min$ objective in a grid environment, where each tile is associated with a deterministic reward sampled uniformly from $[-1, 1]$ when initializing the grid. At each step, the agent can choose to move forward and left, forward and right, or forward and straight, and an episode terminates when the agent has crossed from one side of the grid to the other (see \Cref{fig:reward_map_grid}). We perform experiments on grid sizes $N = {3,4,5}$ with 10 random grids per size. For each grid and training method, we train 5 agents with different initialization. To facilitate a fair comparison between both methods, we use a vanilla deep Q-learning algorithm with minimal hyperparameters for training (for details see \Cref{app:details_experiments}). We either extend the states and adjust the rewards as specified in \Cref{tab:ftable} for the minimum objective and use the standard Q-function update of Q-learning (ours), or we only change the Q-function update as specified by \citet{cui2023} to $Q(s_t, a_t) \leftarrow \min (r_t, \underset{a}{\text{argmax }} Q(s_{t+1}, a))$.
Our results indicate that the overall performance of the two algorithms is similar. While our method may yield a higher final return (see \Cref{tab:cui_vs_us_final_return}), it may train slightly slower (see \Cref{tab:cui_vs_us_auc}).
The prediction loss of the Q-function is also similar for both methods (see \Cref{fig:losses_cui_comp}).

\section{Details on Experiments}\label{app:details_experiments}

\hspace*{\parindent}\textit{Lunar lander:} For training, we use the PPO implementation of \texttt{stables-baselines3} \citep{stable-baselines3}. The hyperparameters and network architecture of both algorithms were chosen as by \citet{rl-zoo3} (which are optimized to give good performance without the velocity penalty), only increasing the batch size and the total training steps to ensure convergence. Training a single agent takes around one hour on a Quadro RTX 6000 GPU with the environment running in parallel on 32 CPUs.

\textit{Portfolio optimization:}
For training, we use the PPO implementation of \texttt{stables-baselines3} \citep{stable-baselines3}.
The hyperparameters and network architecture  of all algorithms were chosen as by \citet{sood2023} where they were optimized to give a good performance of the DIFFSHARPE algorithm. Training a single agent takes around one hour on a Quadro RTX 6000 GPU with the environment running in parallel on 10 CPUs.

\textit{Peak environment:} 
For training, we use a custom REINFORCE implementation with a tabular policy to keep the number of hyperparameters minimal, updating the agent's policy after each completed trajectory. We use a learning rate of $2^{-10}$ but also performed experiments scanning the learning rate which leaves qualitative results similar. Training a single agent takes around 20 minutes on a Quadro RTX 6000 GPU with the environment running on a single CPU.

\textit{ZX-diagrams:} 
For training, we use the PPO implementation of \citet{nagele2023} to facilitate the changing observation and action space.
The hyperparameters and network architecture were chosen as by \citet{nagele2023} who originally chose them to give good performance of the PPO algorithm. This is the most compute-intensive experiment reported in this work with one training run lasting for 12 hours using two Quadro RTX 6000 GPUs and 32 CPUs.

\textit{Quantum error correction} For training, we use the PPO implementation of \texttt{PureJaxRL} \citep{lu2022}. 
The hyperparameters have been chosen to be the optimal ones reported in \citet{zen2024}, which were optimized for performance of the PPO algorithm. All of the training is done on a single Quadro RTX 6000 GPU. 
Training 10 agents to complete the code discovery task takes  1 to 2 minutes, depending on the target code parameters. For the logical state preparation task, it takes around 100 seconds for $[[5,1,3]]_l$ and around 2500 seconds for $[[17,1,5]]_l$ to train 10 agents in parallel.

\textit{Grid environment:}
For training, we use a vanilla Q-learning algorithm. Per training run, we take a total of $10^5$ steps in the environment. After a warmup phase of 1000 steps, we train after every 100 steps taken in the environment on a batch consisting of the last 1000 experiences of the agent. For exploration, we use an $\epsilon$-greedy strategy with $\epsilon=0.1$. We use the Adam optimizer with the learning rate linearly annealed from $10^{-2}$ to $10^{-7}$. For the Q-network, we use a multilayer perceptron with 2 hidden layers of dimension 128 with tanh activations.
One training run takes around 30 seconds on a single CPU.

\textit{Total compute resources:} We estimate the total compute time to reproduce the results of this manuscript to be around 1250 GPU hours and 25000 CPU hours.

\begin{figure}[t]
    \centering
    \includegraphics[width=0.85\textwidth]{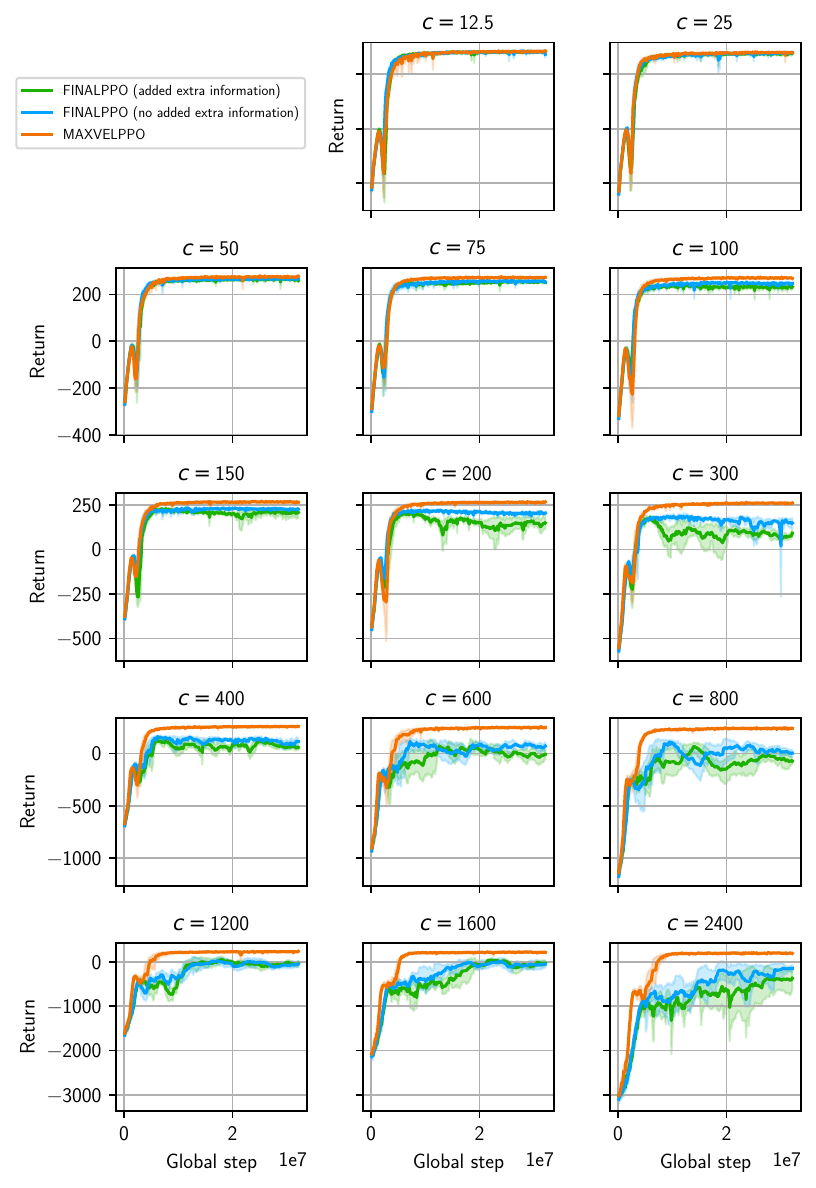}
    \caption{Training progress of the cumulative FINALPPO algorithm (with and without added state information $h_t$) and of the non-cumulative MAXVELPPO algorithm in the lunar lander environment.}
    \label{fig:fig_train_ll}
\end{figure}

\begin{figure}[t]
    \centering
    \includegraphics[width=0.85\textwidth]{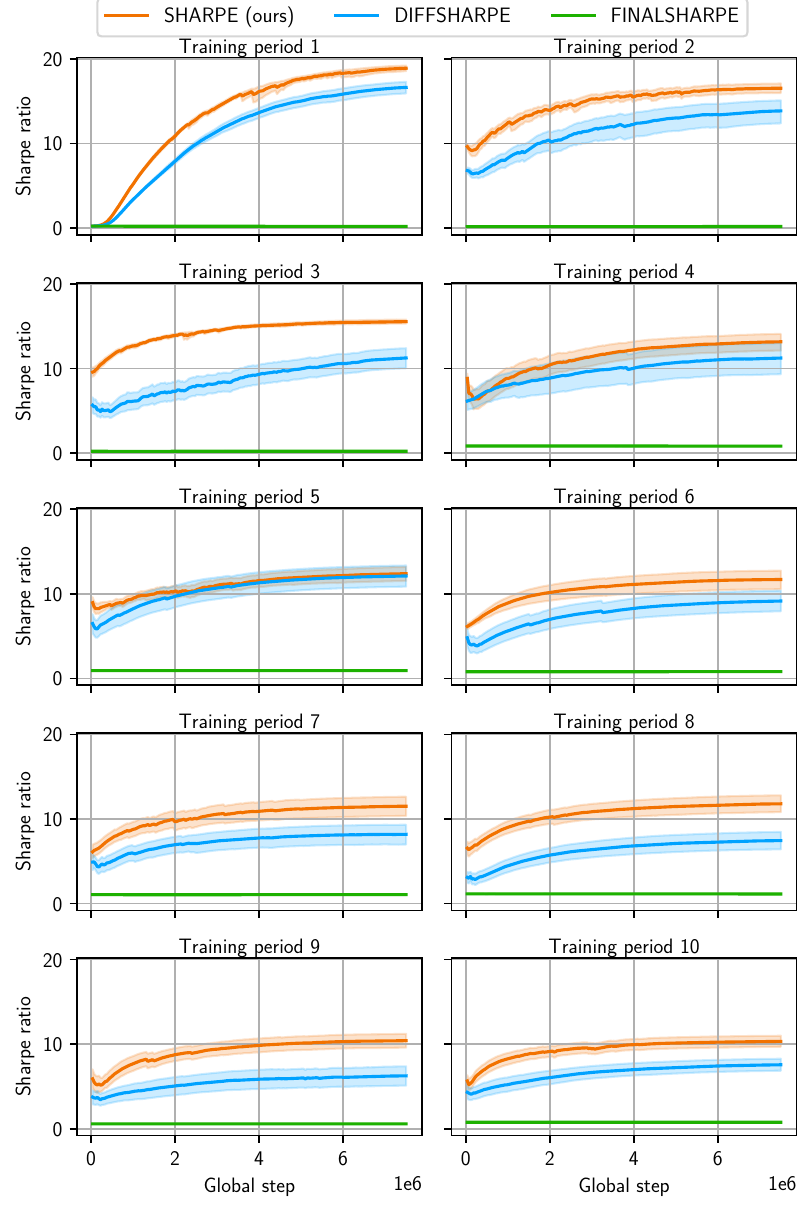}
    \caption{Training progress in the portfolio optimization task.}
    \label{fig:fig_train_finance}
\end{figure}

\begin{figure}[t]
    \centering
    \includegraphics[width=0.85\textwidth]{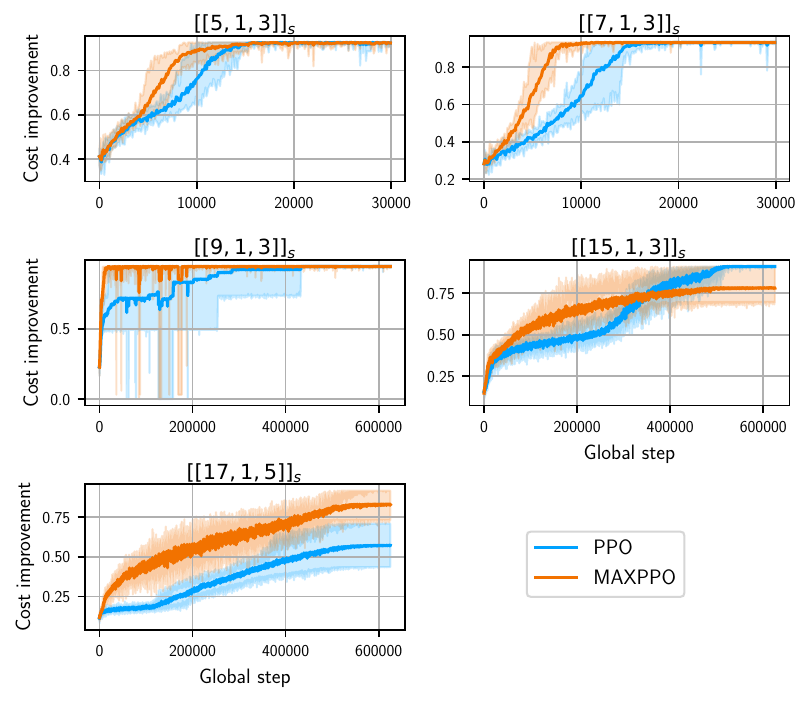}
    \caption{Training progress in the quantum error correction tasks. The best and worst performance of 10 seeds is shaded.}
    \label{fig:fig_train_qec}
\end{figure}

\begin{figure}[h]
    \centering
    \includegraphics{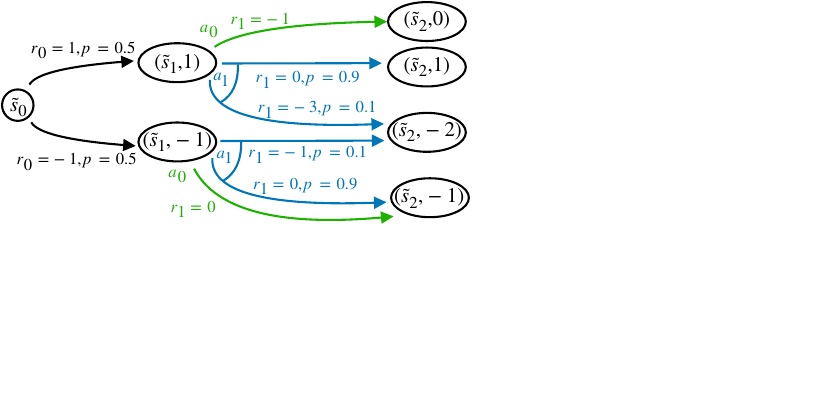}
    \caption{Markov decision process constructed from the non-cumulative Markov decision process depicted in \Cref{fig:fig2}.}
    \label{fig:figA1}
\end{figure}

\begin{figure}[h]
\centering
\begin{minipage}[t]{0.5\textwidth}
  \centering
  \begin{tabular}{ ccc } 
  \toprule
 State $s$ & Action $a$ & $Q(s,a)$\\  \midrule
 $\tilde s_0$ & - & $-0.15$ \\
 $(\tilde s_1, 1)$& $a_0$ & $-1$ \\
 $(\tilde s_1, 1)$& $a_1$ & $-0.3$ \\
 $(\tilde s_1, -1)$ & $a_0$ & $0$ \\
 $(\tilde s_1, -1)$ & $a_1$ & $-0.1$ \\
 \bottomrule
\end{tabular}
\captionsetup{width=0.95\linewidth}
\captionof{table}{Q-function $Q$ of the MDP depicted in \Cref{fig:figA1} (which corresponds to the NCMDP of \Cref{fig:fig2}) found by value iteration using our method. The resulting policy is optimal and results in an expected return
of $-0.15$.}
\label{tab:Qtable}
\end{minipage}%
\begin{minipage}[t]{0.5\textwidth}
  \centering
  \begin{tabular}{ ccc } 
  \toprule
 State $\tilde s$ & Action $a$ & $Q^\prime(\tilde s,a)$\\  \midrule
 $\tilde s_0$ & - & $-0.5$ \\
 $\tilde s_1$& $a_0$ & $0$ \\
 $\tilde s_1$& $a_1$ & $-0.2$ \\
 \bottomrule
\end{tabular}
\captionsetup{width=0.95\linewidth}
\captionof{table}{Q-function $Q^\prime$ of the MDP depicted in \Cref{fig:fig2} found by value iteration using the method of \citet{cui2023}. The resulting policy is not optimal and results in an expected return of $-0.5$.}
\label{tab:QtableCui}
\end{minipage}
\end{figure}

\FloatBarrier


\begin{thebibliography}{43}
\providecommand{\natexlab}[1]{#1}
\providecommand{\url}[1]{\texttt{#1}}
\expandafter\ifx\csname urlstyle\endcsname\relax
  \providecommand{\doi}[1]{doi: #1}\else
  \providecommand{\doi}{doi: \begingroup \urlstyle{rm}\Url}\fi

\bibitem[Andrychowicz et~al.(2020{\natexlab{a}})Andrychowicz, Baker, Chociej,
  Józefowicz, McGrew, Pachocki, Petron, Plappert, Powell, Ray, Schneider,
  Sidor, Tobin, Welinder, Weng, and Zaremba]{OpenAI2020}
Marcin Andrychowicz, Bowen Baker, Maciek Chociej, Rafal Józefowicz, Bob
  McGrew, Jakub Pachocki, Arthur Petron, Matthias Plappert, Glenn Powell, Alex
  Ray, Jonas Schneider, Szymon Sidor, Josh Tobin, Peter Welinder, Lilian Weng,
  and Wojciech Zaremba.
\newblock Learning dexterous in-hand manipulation.
\newblock \emph{The International Journal of Robotics Research}, 39\penalty0
  (1):\penalty0 3--20, 2020{\natexlab{a}}.
\newblock \doi{10.1177/0278364919887447}.
\newblock URL \url{https://journals.sagepub.com/doi/10.1177/0278364919887447}.

\bibitem[Andrychowicz et~al.(2020{\natexlab{b}})Andrychowicz, Raichuk,
  Sta{\'n}czyk, Orsini, Girgin, Marinier, Hussenot, Geist, Pietquin, and
  Michalski]{andrychowicz2020}
Marcin Andrychowicz, Anton Raichuk, Piotr Sta{\'n}czyk, Manu Orsini, Sertan
  Girgin, Rapha{\"e}l Marinier, Leonard Hussenot, Matthieu Geist, Olivier
  Pietquin, and Marcin Michalski.
\newblock What matters for on-policy deep actor-critic methods? {A} large-scale
  study.
\newblock In \emph{International conference on learning representations},
  2020{\natexlab{b}}.
\newblock \doi{10.48550/arXiv.2006.05990}.
\newblock URL \url{https://openreview.net/forum?id=nIAxjsniDzg}.

\bibitem[Coecke and Kissinger(2017)]{Coecke_2017}
Bob Coecke and Aleks Kissinger.
\newblock \emph{Picturing Quantum Processes: A First Course in Quantum Theory
  and Diagrammatic Reasoning}.
\newblock Cambridge University Press, 2017.
\newblock ISBN 9781107104228.

\bibitem[Cui and Yu(2023)]{cui2023}
Wei Cui and Wei Yu.
\newblock Reinforcement learning with non-cumulative objective.
\newblock \emph{IEEE Transactions on Machine Learning in Communications and
  Networking}, 1:\penalty0 124--137, 2023.
\newblock \doi{10.1109/TMLCN.2023.3285543}.
\newblock URL \url{https://ieeexplore.ieee.org/document/10151914}.

\bibitem[Duncan et~al.(2020)Duncan, Kissinger, Perdrix, and Van
  De~Wetering]{duncan2020}
Ross Duncan, Aleks Kissinger, Simon Perdrix, and John Van De~Wetering.
\newblock Graph-theoretic simplification of quantum circuits with the
  {ZX}-calculus.
\newblock \emph{Quantum}, 4:\penalty0 279, 2020.
\newblock \doi{10.22331/q-2020-06-04-279}.
\newblock URL \url{http://quantum-journal.org/papers/q-2020-06-04-279/}.

\bibitem[Eyckerman et~al.(2022)Eyckerman, Reiter, Latré, Marquez-Barja, and
  Hellinckx]{eyckerman2022}
Reinout Eyckerman, Phil Reiter, Steven Latré, Johann Marquez-Barja, and Peter
  Hellinckx.
\newblock Application placement in fog environments using multi-objective
  reinforcement learning with maximum reward formulation.
\newblock In \emph{NOMS 2022-2022 IEEE/IFIP Network Operations and Management
  Symposium}, pages 1--6, 2022.
\newblock \doi{10.1109/NOMS54207.2022.9789757}.
\newblock URL \url{https://ieeexplore.ieee.org/document/9789757}.

\bibitem[Fisac et~al.(2015)Fisac, Chen, Tomlin, and Sastry]{fisac2015}
Jaime~F Fisac, Mo~Chen, Claire~J Tomlin, and S~Shankar Sastry.
\newblock Reach-avoid problems with time-varying dynamics, targets and
  constraints.
\newblock In \emph{Proceedings of the 18th international conference on hybrid
  systems: computation and control}, pages 11--20, 2015.
\newblock \doi{10.1145/2728606.2728612}.
\newblock URL \url{https://dl.acm.org/doi/10.1145/2728606.2728612}.

\bibitem[Fisac et~al.(2019)Fisac, Lugovoy, Rubies-Royo, Ghosh, and
  Tomlin]{fisac2019}
Jaime~F. Fisac, Neil~F. Lugovoy, Vicenç Rubies-Royo, Shromona Ghosh, and
  Claire~J. Tomlin.
\newblock Bridging {Hamilton-Jacobi} safety analysis and reinforcement
  learning.
\newblock In \emph{2019 International Conference on Robotics and Automation
  (ICRA)}, pages 8550--8556, 2019.
\newblock \doi{10.1109/ICRA.2019.8794107}.
\newblock URL \url{https://ieeexplore.ieee.org/document/8794107}.

\bibitem[F{\"o}sel et~al.(2021)F{\"o}sel, Niu, Marquardt, and Li]{fosel2021}
Thomas F{\"o}sel, Murphy~Yuezhen Niu, Florian Marquardt, and Li~Li.
\newblock Quantum circuit optimization with deep reinforcement learning.
\newblock \emph{arXiv}, 2021.
\newblock \doi{10.48550/arXiv.2103.07585}.
\newblock URL \url{https://arxiv.org/abs/2103.07585}.

\bibitem[Gottipati et~al.(2020)Gottipati, Pathak, Nuttall, Chunduru, Touati,
  Subramanian, Taylor, and Chandar]{gottipati2020}
Sai~Krishna Gottipati, Yashaswi Pathak, Rohan Nuttall, Raviteja Chunduru, Ahmed
  Touati, Sriram~Ganapathi Subramanian, Matthew~E Taylor, and Sarath Chandar.
\newblock Maximum reward formulation in reinforcement learning.
\newblock \emph{arXiv}, 2020.
\newblock \doi{10.48550/arXiv.2010.03744}.
\newblock URL \url{https://arxiv.org/abs/2010.03744}.

\bibitem[Hsu et~al.(2021)Hsu, Rubies-Royo, Tomlin, and Fisac]{hsu2021}
Kai-Chieh Hsu, Vicen{\c{c}} Rubies-Royo, Claire~J Tomlin, and Jaime~F Fisac.
\newblock Safety and liveness guarantees through reach-avoid reinforcement
  learning.
\newblock In \emph{Proceedings of Robotics: Science and Systems}, July 2021.
\newblock \doi{10.15607/RSS.2021.XVII.077}.
\newblock URL \url{https://www.roboticsproceedings.org/rss17/p077.pdf}.

\bibitem[Kaledin et~al.(2022)Kaledin, Golubev, and Belomestny]{kaledin2022}
Maxim Kaledin, Alexander Golubev, and Denis Belomestny.
\newblock Variance reduction for policy-gradient methods via empirical variance
  minimization.
\newblock \emph{arXiv}, 2022.
\newblock \doi{10.48550/arXiv.2206.06827}.
\newblock URL \url{https://arxiv.org/abs/2206.06827}.

\bibitem[Li et~al.(2020)Li, Wang, Lin, Sinha, and Wellman]{Li_2020}
Junyi Li, Xintong Wang, Yaoyang Lin, Arunesh Sinha, and Michael Wellman.
\newblock Generating realistic stock market order streams.
\newblock \emph{Proceedings of the AAAI Conference on Artificial Intelligence},
  34:\penalty0 727--734, 2020.
\newblock \doi{10.1609/aaai.v34i01.5415}.
\newblock URL \url{https://ojs.aaai.org/index.php/AAAI/article/view/5415}.

\bibitem[Lu et~al.(2022)Lu, Kuba, Letcher, Metz, Schroeder~de Witt, and
  Foerster]{lu2022}
Chris Lu, Jakub Kuba, Alistair Letcher, Luke Metz, Christian Schroeder~de Witt,
  and Jakob Foerster.
\newblock Discovered policy optimisation.
\newblock \emph{Advances in Neural Information Processing Systems},
  35:\penalty0 16455--16468, 2022.
\newblock \doi{10.48550/arXiv.2210.05639}.
\newblock URL
  \url{https://proceedings.neurips.cc/paper_files/paper/2022/hash/688c7a82e31653e7c256c6c29fd3b438-Abstract-Conference.html}.

\bibitem[Mankowitz et~al.(2023)Mankowitz, Michi, Zhernov, Gelmi, Selvi,
  Paduraru, Leurent, Iqbal, Lespiau, Ahern, K{\"o}ppe, Millikin, Gaffney,
  Elster, Broshear, Gamble, Milan, Tung, Hwang, Cemgil, Barekatain, Li,
  Mandhane, Hubert, Schrittwieser, Hassabis, Kohli, Riedmiller, Vinyals, and
  Silver]{Mankowitz2023}
Daniel~J. Mankowitz, Andrea Michi, Anton Zhernov, Marco Gelmi, Marco Selvi,
  Cosmin Paduraru, Edouard Leurent, Shariq Iqbal, Jean-Baptiste Lespiau, Alex
  Ahern, Thomas K{\"o}ppe, Kevin Millikin, Stephen Gaffney, Sophie Elster,
  Jackson Broshear, Chris Gamble, Kieran Milan, Robert Tung, Minjae Hwang,
  Taylan Cemgil, Mohammadamin Barekatain, Yujia Li, Amol Mandhane, Thomas
  Hubert, Julian Schrittwieser, Demis Hassabis, Pushmeet Kohli, Martin
  Riedmiller, Oriol Vinyals, and David Silver.
\newblock Faster sorting algorithms discovered using deep reinforcement
  learning.
\newblock \emph{Nature}, 618\penalty0 (7964):\penalty0 257--263, 2023.
\newblock \doi{10.1038/s41586-023-06004-9}.
\newblock URL \url{https://doi.org/10.1038/s41586-023-06004-9}.

\bibitem[Mnih et~al.(2015)Mnih, Kavukcuoglu, Silver, Rusu, Veness, Bellemare,
  Graves, Riedmiller, Fidjeland, Ostrovski, Petersen, Beattie, Sadik,
  Antonoglou, King, Kumaran, Wierstra, Legg, and Hassabis]{Mnih2015}
Volodymyr Mnih, Koray Kavukcuoglu, David Silver, Andrei~A. Rusu, Joel Veness,
  Marc~G. Bellemare, Alex Graves, Martin Riedmiller, Andreas~K. Fidjeland,
  Georg Ostrovski, Stig Petersen, Charles Beattie, Amir Sadik, Ioannis
  Antonoglou, Helen King, Dharshan Kumaran, Daan Wierstra, Shane Legg, and
  Demis Hassabis.
\newblock Human-level control through deep reinforcement learning.
\newblock \emph{Nature}, 518\penalty0 (7540):\penalty0 529--533, 2015.
\newblock \doi{10.1038/nature14236}.
\newblock URL \url{https://doi.org/10.1038/nature14236}.

\bibitem[Moflic and Paler(2023)]{moflic2023}
Ioana Moflic and Alexandru Paler.
\newblock Cost explosion for efficient reinforcement learning optimisation of
  quantum circuits.
\newblock In \emph{2023 IEEE International Conference on Rebooting Computing
  (ICRC)}, pages 1--5. IEEE, 2023.
\newblock \doi{10.1109/ICRC60800.2023.10386864}.
\newblock URL
  \url{https://www.computer.org/csdl/proceedings-article/icrc/2023/10386864/1TJmieJCklW}.

\bibitem[Moody and Saffell(2001)]{moody2001}
J.~Moody and M.~Saffell.
\newblock Learning to trade via direct reinforcement.
\newblock \emph{IEEE Transactions on Neural Networks}, 12\penalty0
  (4):\penalty0 875--889, 2001.
\newblock \doi{10.1109/72.935097}.
\newblock URL \url{https://ieeexplore.ieee.org/document/935097}.

\bibitem[Moody et~al.(1998)Moody, Wu, Liao, and Saffell]{moody1998}
John Moody, Lizhong Wu, Yuansong Liao, and Matthew Saffell.
\newblock Performance functions and reinforcement learning for trading systems
  and portfolios.
\newblock \emph{Journal of Forecasting}, 17\penalty0 (5-6):\penalty0 441--470,
  1998.
\newblock
  \doi{https://doi.org/10.1002/(SICI)1099-131X(1998090)17:5/6<441::AID-FOR707>3.0.CO;2-\#}.
\newblock URL
  \url{https://onlinelibrary.wiley.com/doi/10.1002/%28SICI%291099-131X%281998090%2917%3A5/6%3C441%3A%3AAID-FOR707%3E3.0.CO%3B2-%23}.

\bibitem[N\"agele(2024{\natexlab{a}})]{githubncmdp}
Maximilian N\"agele.
\newblock Code for classical control, portfolio optimization, and peak
  environment.
\newblock \url{https://github.com/MaxNaeg/ncmdp}, 2024{\natexlab{a}}.

\bibitem[N\"agele(2024{\natexlab{b}})]{githubzx}
Maximilian N\"agele.
\newblock Code for {ZX}-diagrams.
\newblock \url{https://github.com/MaxNaeg/ZXreinforce}, 2024{\natexlab{b}}.

\bibitem[N{\"a}gele and Marquardt(2024)]{nagele2023}
Maximilian N{\"a}gele and Florian Marquardt.
\newblock Optimizing {ZX}-diagrams with deep reinforcement learning.
\newblock \emph{Machine Learning: Science and Technology}, 5\penalty0
  (3):\penalty0 035077, 2024.
\newblock \doi{10.1088/2632-2153/ad76f7}.
\newblock URL
  \url{https://iopscience.iop.org/article/10.1088/2632-2153/ad76f7}.

\bibitem[Olle(2024)]{githubqec}
Jan Olle.
\newblock Code for quantum error correction code discovery.
\newblock \url{https://github.com/jolle-ag/qdx}, 2024.

\bibitem[Olle et~al.(2024)Olle, Zen, Puviani, and Marquardt]{olle2023}
Jan Olle, Remmy Zen, Matteo Puviani, and Florian Marquardt.
\newblock Simultaneous discovery of quantum error correction codes and encoders
  with a noise-aware reinforcement learning agent.
\newblock \emph{npj Quantum Information}, 10\penalty0 (1):\penalty0 126, 2024.
\newblock \doi{10.1038/s41534-024-00920-y}.
\newblock URL \url{https://doi.org/10.1038/s41534-024-00920-y}.

\bibitem[Quah and Quek(2006)]{quah2006}
K.H. Quah and Chai Quek.
\newblock Maximum reward reinforcement learning: A non-cumulative reward
  criterion.
\newblock \emph{Expert Systems with Applications}, 31\penalty0 (2):\penalty0
  351--359, 2006.
\newblock \doi{10.1016/j.eswa.2005.09.054}.
\newblock URL
  \url{https://www.sciencedirect.com/science/article/pii/S0957417405002228}.

\bibitem[Raffin(2020)]{rl-zoo3}
Antonin Raffin.
\newblock Rl baselines3 zoo.
\newblock \url{https://github.com/DLR-RM/rl-baselines3-zoo}, 2020.

\bibitem[Raffin et~al.(2021)Raffin, Hill, Gleave, Kanervisto, Ernestus, and
  Dormann]{stable-baselines3}
Antonin Raffin, Ashley Hill, Adam Gleave, Anssi Kanervisto, Maximilian
  Ernestus, and Noah Dormann.
\newblock Stable-baselines3: Reliable reinforcement learning implementations.
\newblock \emph{Journal of Machine Learning Research}, 22\penalty0
  (268):\penalty0 1--8, 2021.
\newblock URL \url{http://jmlr.org/papers/v22/20-1364.html}.

\bibitem[Riu et~al.(2023)Riu, Nogu{\'e}, Vilaplana, Garcia-Saez, and
  Estarellas]{riu2023}
Jordi Riu, Jan Nogu{\'e}, Gerard Vilaplana, Artur Garcia-Saez, and Marta~P
  Estarellas.
\newblock Reinforcement learning based quantum circuit optimization via
  {ZX}-calculus.
\newblock \emph{arXiv}, 2023.
\newblock \doi{10.48550/arXiv.2312.11597}.
\newblock URL \url{https://arxiv.org/abs/2312.11597}.

\bibitem[Schulman et~al.(2017)Schulman, Wolski, Dhariwal, Radford, and
  Klimov]{schulman2017}
John Schulman, Filip Wolski, Prafulla Dhariwal, Alec Radford, and Oleg Klimov.
\newblock Proximal policy optimization algorithms.
\newblock \emph{arXiv}, 2017.
\newblock \doi{10.48550/arXiv.1707.06347}.
\newblock URL \url{https://arxiv.org/abs/1707.06347}.

\bibitem[Sharpe(1966)]{sharpe1966}
William~F. Sharpe.
\newblock Mutual fund performance.
\newblock \emph{The Journal of Business}, 39\penalty0 (1):\penalty0 119--138,
  1966.
\newblock ISSN 00219398, 15375374.
\newblock URL \url{http://www.jstor.org/stable/2351741}.

\bibitem[Sood et~al.(2023)Sood, Papasotiriou, Vaiciulis, and Balch]{sood2023}
Srijan Sood, Kassiani Papasotiriou, Marius Vaiciulis, and Tucker Balch.
\newblock Deep reinforcement learning for optimal portfolio allocation: A
  comparative study with mean-variance optimization.
\newblock \emph{International Conference on Automated Planning and Scheduling},
  page~21, 2023.
\newblock URL
  \url{https://icaps23.icaps-conference.org/papers/finplan/FinPlan23_paper_4.pdf}.

\bibitem[Sutton and Barto(2018)]{sutton1998introduction}
Richard~S. Sutton and Andrew~G. Barto.
\newblock \emph{Reinforcement Learning: {A}n Introduction, Second edition}.
\newblock The MIT Press, 2018.
\newblock ISBN 9780262039246.
\newblock URL \url{http://incompleteideas.net/book/RLbook2020.pdf}.

\bibitem[Sutton et~al.(1999)Sutton, McAllester, Singh, and Mansour]{sutton1999}
Richard~S Sutton, David McAllester, Satinder Singh, and Yishay Mansour.
\newblock Policy gradient methods for reinforcement learning with function
  approximation.
\newblock \emph{Advances in neural information processing systems}, 12, 1999.
\newblock URL
  \url{https://papers.nips.cc/paper_files/paper/1999/hash/464d828b85b0bed98e80ade0a5c43b0f-Abstract.html}.

\bibitem[Terhal(2015)]{Terhal2015}
Barbara~M. Terhal.
\newblock Quantum error correction for quantum memories.
\newblock \emph{Rev. Mod. Phys.}, 87:\penalty0 307--346, 2015.
\newblock \doi{10.1103/RevModPhys.87.307}.
\newblock URL \url{https://link.aps.org/doi/10.1103/RevModPhys.87.307}.

\bibitem[Towers et~al.(2023)Towers, Terry, Kwiatkowski, Balis, Cola, Deleu,
  Goulão, Kallinteris, KG, Krimmel, Perez-Vicente, Pierré, Schulhoff, Tai,
  Shen, and Younis]{towers_gymnasium_2023}
Mark Towers, Jordan~K. Terry, Ariel Kwiatkowski, John~U. Balis, Gianluca~de
  Cola, Tristan Deleu, Manuel Goulão, Andreas Kallinteris, Arjun KG, Markus
  Krimmel, Rodrigo Perez-Vicente, Andrea Pierré, Sander Schulhoff, Jun~Jet
  Tai, Andrew Tan~Jin Shen, and Omar~G. Younis.
\newblock Gymnasium, 2023.
\newblock URL \url{https://zenodo.org/record/8127025}.

\bibitem[Veviurko et~al.(2024)Veviurko, Boehmer, and de~Weerdt]{veviurko2024}
Grigorii Veviurko, Wendelin Boehmer, and Mathijs de~Weerdt.
\newblock To the max: Reinventing reward in reinforcement learning.
\newblock In \emph{41st International Conference on Machine Learning (ICML)},
  2024.
\newblock \doi{10.48550/arXiv.2402.01361}.
\newblock URL \url{https://openreview.net/forum?id=4KQ0VwqPg8}.

\bibitem[Wang et~al.(2020)Wang, Zhong, Du, Salakhutdinov, and Yang]{Wang2020}
Ruosong Wang, Peilin Zhong, Simon~S Du, Russ~R Salakhutdinov, and Lin Yang.
\newblock Planning with general objective functions: Going beyond total
  rewards.
\newblock In H.~Larochelle, M.~Ranzato, R.~Hadsell, M.F. Balcan, and H.~Lin,
  editors, \emph{Advances in Neural Information Processing Systems}, volume~33,
  pages 14486--14497, 2020.
\newblock URL
  \url{https://proceedings.neurips.cc/paper_files/paper/2020/file/a6a767bbb2e3513233f942e0ff24272c-Paper.pdf}.

\bibitem[Watkins and Dayan(1992)]{watkins1992}
Christopher~JCH Watkins and Peter Dayan.
\newblock Q-learning.
\newblock \emph{Machine learning}, 8:\penalty0 279--292, 1992.
\newblock \doi{10.1007/BF00992698}.
\newblock URL \url{https://link.springer.com/article/10.1007/BF00992698}.

\bibitem[Williams(1992)]{williams1992}
Ronald~J. Williams.
\newblock Simple statistical gradient-following algorithms for connectionist
  reinforcement learning.
\newblock \emph{Machine Learning}, 8\penalty0 (3):\penalty0 229--256, 1992.
\newblock \doi{10.1007/BF00992696}.
\newblock URL \url{https://doi.org/10.1007/BF00992696}.

\bibitem[Yu et~al.(2022)Yu, Ma, Li, and Chen]{yu2022}
Dongjie Yu, Haitong Ma, Shengbo Li, and Jianyu Chen.
\newblock Reachability constrained reinforcement learning.
\newblock In \emph{Proceedings of the 39th International Conference on Machine
  Learning}, volume 162, pages 25636--25655, 2022.
\newblock \doi{10.48550/arXiv.2205.07536}.
\newblock URL \url{https://proceedings.mlr.press/v162/yu22d.html}.

\bibitem[Zen(2024)]{githubls}
Remmy Zen.
\newblock Code for logical state preparation.
\newblock \url{https://github.com/remmyzen/rlftqc}, 2024.

\bibitem[Zen et~al.(2024)Zen, Olle, Colmenarez, Puviani, M{\"u}ller, and
  Marquardt]{zen2024}
Remmy Zen, Jan Olle, Luis Colmenarez, Matteo Puviani, Markus M{\"u}ller, and
  Florian Marquardt.
\newblock Quantum circuit discovery for fault-tolerant logical state
  preparation with reinforcement learning.
\newblock \emph{arXiv}, 2024.
\newblock \doi{10.48550/arXiv.2402.1776}.
\newblock URL \url{https://arxiv.org/abs/2402.17761}.

\bibitem[Zhou et~al.(2019)Zhou, Kearnes, Li, Zare, and Riley]{zhou2019}
Zhenpeng Zhou, Steven Kearnes, Li~Li, Richard~N. Zare, and Patrick Riley.
\newblock Optimization of molecules via deep reinforcement learning.
\newblock \emph{Scientific Reports}, 9:\penalty0 10752, 2019.
\newblock \doi{10.1038/s41598-019-47148-x}.
\newblock URL \url{https://www.nature.com/articles/s41598-019-47148-x}.

\end{thebibliography}

\end{appendices}

\end{document}